\documentclass{article}
\usepackage[main,final]{neurips_2026}

\usepackage{amsmath}
\usepackage{amssymb}
\usepackage{booktabs}
\usepackage{graphicx}
\usepackage{algorithm}
\usepackage{algorithmic}
\usepackage{hyperref}
\usepackage{url}
\usepackage{xcolor}
\usepackage{amsthm}

\usepackage{arydshln}
\usepackage{wrapfig}
\usepackage{multirow}
\usepackage{booktabs}
\usepackage{longtable}
\usepackage{listings}
\usepackage{subcaption}
\usepackage{array}
\definecolor{kwgreen}{rgb}{0.0,0.5,0.0}
\definecolor{strgreen}{rgb}{0.0,0.5,0.0}
\definecolor{commentgray}{rgb}{0.4,0.4,0.4}
\definecolor{backcolour}{rgb}{0.95,0.95,0.95}

\lstdefinestyle{iostyle}{
    language=Python,
    basicstyle=\ttfamily\small,
    keywordstyle=\color{kwgreen}\bfseries,
    stringstyle=\color{strgreen},
    commentstyle=\color{commentgray},
    backgroundcolor=\color{white},
    frame=none,
    backgroundcolor=\color{backcolour},
    aboveskip=0.5\baselineskip,
    belowskip=0.5\baselineskip,
    lineskip=-0.5pt,
    showstringspaces=false,
    breaklines=true,
    tabsize=4,
    keepspaces=true,
    columns=fullflexible,
    xleftmargin=0pt,
    linewidth=\textwidth,
}
\title{Agentic Reverse Engineering: \\From I/O to Code with Discovery Agent}

\author{Yihong Dong$^{1,2}$, Jiaru Qian$^{3}$, Haoran Zhang$^{4}$, Peixu Wang$^{5}$, Binhua Li$^{2}$, Zhi Jin$^{3}$\\ \textbf{Yongbin Li$^{2}$, Ge Li$^{3}$, Xiaokang Yang$^{1}$, Xue Jiang$^{2,3}$}\\
$^1$ Shanghai Jiao Tong University \quad $^2$ Tongyi Lab, Alibaba Group \quad \\ $^3$ Peking University \quad $^4$ Renmin University of China \quad $^5$ National University of Singapore
}

\begin{document}

\maketitle

\begin{abstract}
The automatic synthesis of a program from any form of specification is regarded as a holy grail of computer science. Fueled by LLMs, NL2Code has achieved tremendous success, yet the fundamentally more challenging task of synthesizing programs from input-output behavior, which we refer to as IO2Code, remains largely unsolved. Whereas NL2Code can exploit the semantic alignment between natural language and code acquired during pretraining, IO2Code requires recovering underlying principles from concrete computational behavior, navigating a vast and underspecified hypothesis space. To address this, we propose DIO-Agent, a discovery agent that performs agentic reverse engineering from input-output behavior to code.
Our method frames IO2Code as an evolutionary search over discrete program space, in which an LLM serves as the mutation operator and concrete error signals from execution guide each mutation. To prevent the search from wandering into structurally complex yet incorrect dead ends, we introduce the Transformation Priority Premise as a mutation prior that biases the LLM toward the simplest hypothesis consistent with current evidence, progressively escalating from constants to conditionals to iteration only when simpler constructs are insufficient.
To facilitate systematic study, we further construct an IO2CodeBench spanning multiple difficulty levels. Extensive experiments show that DIO-Agent consistently outperforms both traditional program-by-example method and SOTA evolution-agent baselines across all difficulty levels and various LLMs, while substantially surpassing test-time scaling strategies with equivalent budgets.\footnote{Our code and dataset are available at \url{https://github.com/JiaruQian/IO2Code}.}
\end{abstract}

\section{Introduction}
\label{sec:introduction}
\begin{figure*}[t]
    \centering
    \includegraphics[width=0.9\textwidth]{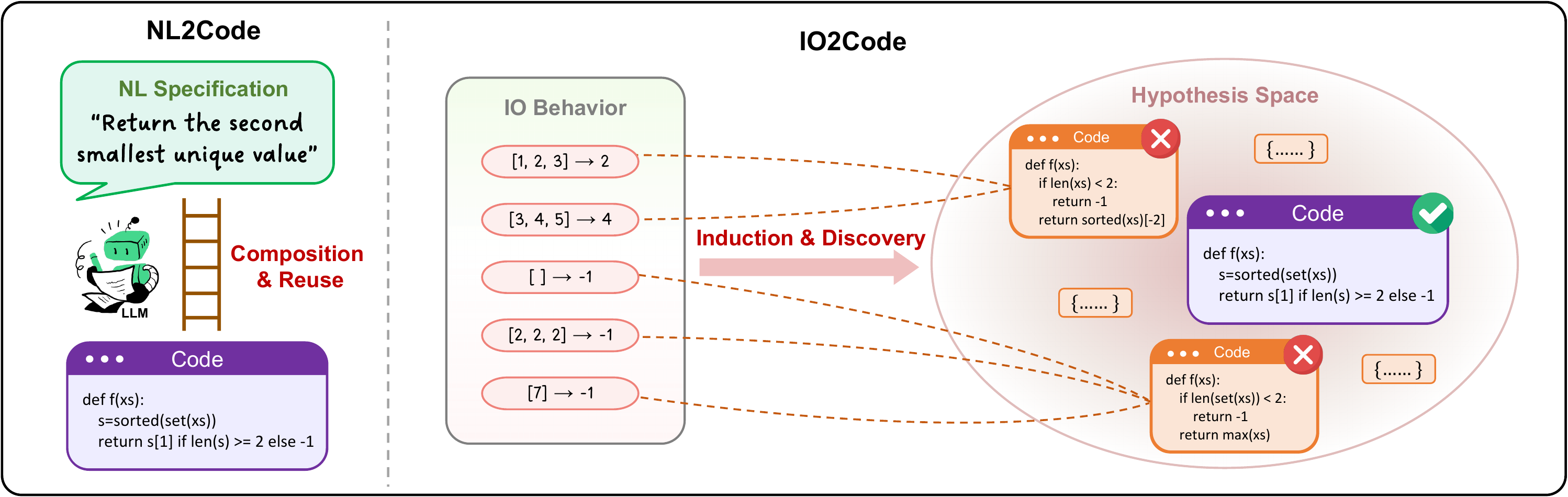}
    \caption{Illustration of IO2Code, where the mainstream NL2Code task evaluates the composition and reuse ability of LLMs, while the IO2Code task evaluates the induction and discovery ability.}
    \label{fig:intro}
\end{figure*}

Large language models (LLMs) have achieved remarkable success in code generation~\citep{chen2021evaluatinglargelanguagemodels, nijkamp2023codegenopenlargelanguage, selfplanning, Self-Collaboration}, largely built upon the NL2Code formulation~\citep{iyer2018mappinglanguagecodeprogrammatic, lu2021codexgluemachinelearningbenchmark,austin2021programsynthesislargelanguage} that converts natural-language descriptions into executable programs~\citep{hendrycks2021measuringcodingchallengecompetence}. NL2Code relies on human-provided descriptions, limiting its applicability when such descriptions are unavailable or unreliable.
However, a wide range of real-world tasks requires developers and researchers to work not from NL descriptions, but from computational behavior. From migrating legacy systems with lost source code and reverse-engineering black-box APIs, to abducing algorithms from scientific observations and capturing user intent through input-output demonstrations, these scenarios span system evolution, scientific discovery, and human-computer interaction~\citep{Collie_2020}. 
We refer to this class of tasks as IO2Code, where the goal is to synthesize programs from input-output behavior (Figure~\ref{fig:intro}).

IO2Code poses fundamentally more severe challenges than NL2Code. NL2Code largely reduces to composing and reusing code patterns aligned with natural-language descriptions learned during pretraining~\citep{zan2023largelanguagemodelsmeet,feng2020codebertpretrainedmodelprogramming,wang2021codet5identifierawareunifiedpretrained}. In contrast, IO2Code demands induction and discovery, where the model must uncover latent algorithmic structures from concrete computational behavior, infer causal relationships between inputs and outputs, and autonomously construct complete program logic. This requires abductive reasoning from the specific to the general, in which the composition and reuse strategies effective in NL2Code become inapplicable. Furthermore, input-output observations are inherently underspecified, as multiple programs can be consistent with a finite set of examples, demanding that the agent navigate a vast hypothesis space while avoiding overfitting to spurious patterns.

To tackle the challenge of IO2Code, we draw on a well-established observation from software engineering. In the practice of Test-Driven Development (TDD), the Transformation Priority Premise (TPP)~\citep{martin2021clean} shows that programmers construct algorithms that pass all tests most effectively when they apply transformations in order of increasing complexity, progressing from constants to variables, from unconditional statements to conditionals, and from conditionals to loops or recursion, and that violating this ordering frequently leads to impasses requiring costly rewrites. We observe that TPP can be repurposed as a structural constraint on autonomous program discovery. In IO2Code, where no specification exists beyond input-output behaviors, such a complexity ordering offers a principled mechanism to force the search process to exhaust simpler hypotheses before entertaining structurally more complex ones, thereby preventing both overfitting to surface patterns and aimless wandering in the program space.

In this paper, we propose DIO-Agent, a discovery agent for IO2Code. Our method formulates the task as an evolutionary search in which an LLM acts as the mutation operator. In the discrete program space without natural-language guidance, allowing a model to mutate code blindly can easily lead it into structurally complex yet incorrect dead ends. To address this, we introduce TPP as a mutation prior that biases the search toward the simplest hypothesis first, escalating to more expressive constructs only when simpler ones fail to explain the observed behavior. This evolutionary process is organized as a curriculum of increasing difficulty, with concrete error signals from execution grounding each mutation in debugging evidence rather than scalar rewards. To facilitate research, we establish an IO2CodeBench spanning multiple difficulty levels from basic data operations to complex algorithmic reasoning.

We conduct extensive experiments on the IO2CodeBench across all difficulty levels. The results demonstrate that DIO-Agent achieves consistent improvements over existing approaches, including traditional PBE methods and SOTA evolution-agent baselines, while substantially surpassing test-time scaling strategies with equivalent sampling budgets. Moreover, DIO-Agent yields stable performance gains across diverse base models, confirming its generalizability. We further validate the effectiveness of each component through comprehensive ablation studies, hyperparameter sensitivity analyses, and illustrative case studies.

\section{Motivation Example}
\label{sec:motivation}

To illustrate the challenges of IO2Code and motivate our approach, consider a concrete example. Suppose an agent is presented with the following input-output pairs and asked to synthesize a program \texttt{f} consistent with all of them:
\begin{center}
\small
\begin{tabular}{llll}
\texttt{f(1) = []} & \texttt{f(2) = [2]} & \texttt{f(3) = [3]} & \texttt{f(4) = [2, 2]} \\
\texttt{f(6) = [2, 3]} & \texttt{f(8) = [2, 2, 2]} & \texttt{f(12) = [2, 2, 3]} & \texttt{f(30) = [2, 3, 5]} \\
\end{tabular}
\end{center}
\noindent To human with mathematical training, the underlying rule is recognizable: \texttt{f} computes prime factorization of its input. But the agent receives no such hint. It has no access to the function name, docstring, or NL description, and must recover the rule purely from computational behavior.

\paragraph{The Overfitting Trap.}

The most straightforward strategy, and the one that unconstrained models tend to adopt, is to memorize the observations directly:
\begin{lstlisting}[language=Python]
def f(n):
    if n == 1: return []
    elif n == 2: return [2]
    elif n == 3: return [3]
    ...
\end{lstlisting}
\noindent This program mirrors the tests rather than generalizing from them. It is a lookup table, not an algorithm. It will fail on any input not seen during synthesis such as \texttt{f(9)} or \texttt{f(100)}. In NL2Code, such overfitting is unlikely because the NL specification (e.g., ``compute the prime factors of~\texttt{n}'') already constrains the model toward a general solution. In IO2Code, no such constraint exists, and the agent faces an underspecified hypothesis space in which memorization is always locally optimal.

\paragraph{The Blind Mutation Trap.}

An alternative strategy is to apply evolutionary search, randomly mutating candidate programs and selecting those that pass more test cases. However, without structural guidance, this search quickly encounters a different failure mode. The agent may produce programs that are syntactically complex yet semantically wrong. For example, a program that computes divisors rather than prime factors, or one that produces outputs of the correct length but with incorrect values. These candidates occupy complex regions of the program space that are difficult to escape through further mutation, creating structural dead ends. The combinatorial explosion of the unconstrained discrete program space makes blind search intractable for all but the simplest tasks.

\paragraph{From Specific to General.}

Both failure modes share a root cause: the absence of a strategy for navigating the hypothesis space from simple to complex. TPP~\citep{martin2021clean} provides exactly such a strategy. In the context of test-driven development, correct algorithms emerge most reliably when programmers apply behavior-changing transformations in order of increasing structural complexity. TPP identifies a priority ordering that progresses from returning constants to variables, to conditionals, to loops, and ultimately to recursion, and demonstrates that adhering to this ordering avoids the impasses that arise when complex transformations are attempted prematurely.
To see how such a principle might apply to IO2Code, consider what happens when the agent is required to follow this complexity ordering on our running example, progressing through levels of increasing structural complexity and exhausting each level before advancing to the next:

\paragraph{Levels 1--2: Constant and Variable.}
The agent begins with the simplest possible programs:
\noindent
\begin{minipage}[t]{0.48\textwidth}
\begin{lstlisting}[language=Python]
def f(n):
    return []
\end{lstlisting}
\end{minipage}%
\hfill
\begin{minipage}[t]{0.48\textwidth}
\begin{lstlisting}[language=Python]
def f(n):
    return [n]
\end{lstlisting}
\end{minipage}
\noindent The constant \texttt{return []} handles only \texttt{f(1)}. Generalizing to \texttt{return [n]} covers \texttt{f(2)} and \texttt{f(3)}, but regresses on \texttt{f(1) = []} and fails on \texttt{f(4) = [2, 2]}. No variable-level program can satisfy all observations. This signals that the rule involves decomposing~\texttt{n}, not merely returning it.

\paragraph{Level 3: Conditional.}
The agent introduces a conditional to split the execution path:
\begin{lstlisting}[language=Python]
def f(n):
    factors = []
    if n % 2 == 0:
        factors.append(2)
        n = n // 2
    if n > 1:
        factors.append(n)
    return factors
\end{lstlisting}
\noindent This captures the idea of divisibility testing but only applies it once. The case \texttt{f(8) = [2, 2, 2]} reveals that a single conditional is insufficient: the operation must be repeated.

\paragraph{Level 4: Iteration.}
The agent transforms the inner \texttt{if} into a \texttt{while}:
\begin{lstlisting}[language=Python]
def f(n):
    factors = []
    d = 2
    while d <= n:
        while n % d == 0:
            factors.append(d)
            n = n // d
        d += 1
    return factors
\end{lstlisting}
\noindent The transformation from \texttt{if} to \texttt{while} allows the resulting program to generalize to all inputs, including those never seen during synthesis.

The strategy provides three benefits for IO2Code. It prevents overfitting by forcing the agent to exhaust simpler hypotheses before resorting to case-specific logic, structures the search space into a sequence of nested and increasingly expressive subspaces, and produces informative failure patterns at each stage that reveal what additional structural complexity is needed, guiding the agent toward the next level of generalization.

\section{Methodology}
\label{sec:method}

In this section, we first formalize the \textbf{IO2Code} task, and then describe the overview of \textbf{DIO-Agent}, finally introduce its key elements in detail, including \textbf{curriculum-wise evolution}, \textbf{transformation-priority guided mutation}, and \textbf{error-grounded feedback}. 

\paragraph{Formalization of IO2Code.}
The traditional code generation task typically relies on natural language intents. In contrast, the IO2Code problem aims to synthesize a generalized program strictly from a limited set of Input-Output (I/O) pairs.  Formally,  an IO2Code task is defined by a set of visible examples $\mathcal{D}_{\text{vis}} = \{(x_i, y_i)\}_{i=1}^{n}$, where each $x_i$ is an input (scalar, list, tuple, or nested structure) and $y_i$ is the corresponding output produced by an unknown target program $p^*$. The objective is to synthesize a program $\hat{p}$ such that $\hat{p}(x_i) = y_i$ for all visible examples and, critically, $\hat{p}$ generalizes to held-out test cases $\mathcal{D}_{\text{test}} = \{(x_j, y_j)\}_{j=1}^{m}$ that the agent never observes during search. Fitness on the visible set is defined as:
\begin{equation}
f(\hat{p}) = \frac{1}{n} \sum_{i=1}^{n} \mathbf{1}[\hat{p}(x_i) = y_i],
\end{equation}
where $\mathbf{1}[\cdot]$ is the indicator function. A candidate is considered a \emph{solution} if and only if $f(\hat{p}) = 1.0$ on the visible set and it additionally passes all held-out tests---the latter serving as a generalization check against overfitting to spurious patterns in the visible examples.

\paragraph{DIO-Agent Backbone.}

DIO-Agent addresses the IO2Code challenge with an LLM-driven evolutionary loop in which LLMs act as mutation operators and executable evaluators provide the selection signal. Search starts from an initial program template copied to multiple islands so that different search trajectories can unfold in parallel.

Within each island, DIO-Agent repeatedly performs four steps: (1) sampling a parent via a mixture of random exploration, fitness-oriented exploitation, and weighted sampling; (2) building a compact prompt context containing the parent, the two best-performing programs from the same island, and one randomly sampled inspiration program; (3) asking the LLM to propose a \texttt{SEARCH/REPLACE} diff rather than a full rewrite, preserving already working code; and (4) executing the child in a sandbox, scoring it, and inserting it back if novel and competitive. Periodic migration exchanges useful programs across islands to avoid premature convergence.

\begin{figure}[t]
    \centering
    \includegraphics[width=1.0\textwidth]{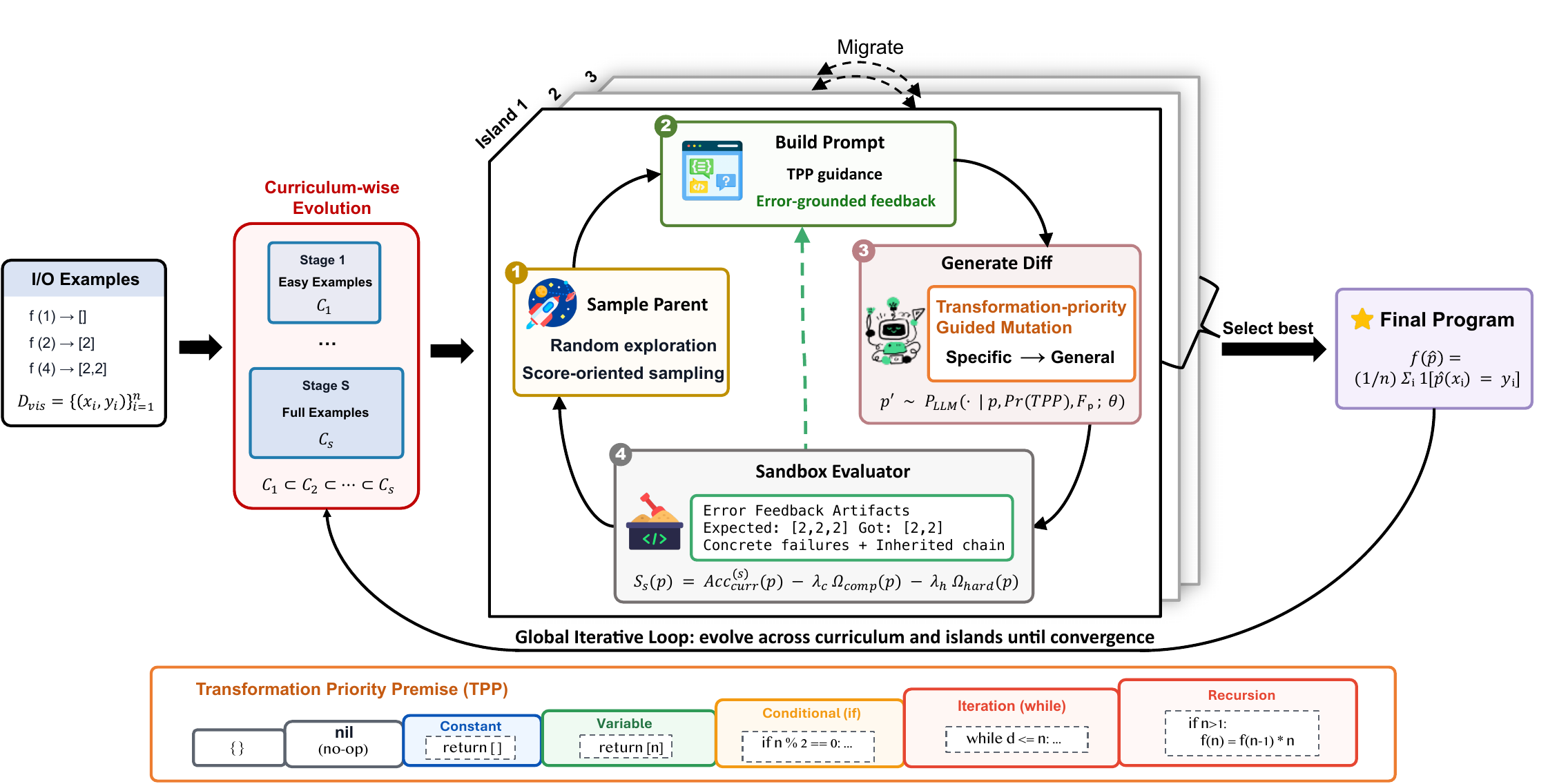}
    \caption{Overview of DIO-Agent. 
    }
    \label{fig:framework}
\end{figure}

This loop is specialized through three tightly coupled design choices: Curriculum-wise Evolution organizes the process stage by stage; Transformation Priority Premise shapes how mutations are proposed within each stage; and Error-Grounded Feedback turns execution results into structured feedback and adopts a fine-grained score to guide evolution. The best program of each stage is passed on as the initial program for the next. The overview of DIO-Agent is illustrated in Figure~\ref{fig:framework}.

\subsection{Curriculum-wise Evolution}

DIO-Agent organizes the evolving process stage by stage. We first sort visible examples from easy to hard using the heuristic that favors shorter and structurally simpler inputs and outputs. This yields a nested curriculum:
\begin{equation}
\mathcal{C}_1 \subset \mathcal{C}_2 \subset \cdots \subset \mathcal{C}_S=\mathcal{D}_{\mathrm{vis}},
\end{equation}
where each stage reveals only a prefix of the full visible set. Let $\Delta_s=\mathcal{C}_s\setminus\mathcal{C}_{s-1}$ denote the newly introduced examples at stage $s$.

Stage $1$ starts from the initial program template of the task. For every later stage $s>1$, DIO-Agent initializes all islands with the best program $p_{s-1}^{*}$ found in stage $s-1$, rather than restarting search from scratch. Inside stage $s$, the mutation prompt only exposes examples in $\mathcal{C}_s$ and explicitly highlights $\Delta_s$, while the evaluator additionally samples $\mathcal{R}_s\subseteq \mathcal{C}_{s-1}$ as replay cases. 

This stage-wise decomposition changes both \emph{what} the model must solve and \emph{where} evolution begins. Early stages encourage simple, high-bias rules that solve easy examples; later stages broaden the behavioral specification and refine an already partially correct program. The final stage exposes all visible examples, and the resulting best program $p_S^{*}$ is then evaluated once on $\mathcal{D}_{\mathrm{test}}$.

\subsection{Transformation-priority Guided Mutation}

Inspired by Test-Driven Development (TDD), we introduce the Transformation Priority Premise (TPP). In a vast, non-convex code space, uniform stochastic prompting invariably leads to semantic degradation. Furthermore, evaluating a program solely by its correctness is insufficient for induction and discovery. DIO-Agent injects TPP into every mutation.

Concretely, the evolution presents an ordered preference over progressively more expressive transformations as follows: \texttt{nil$\rightarrow$constant}, \texttt{constant$\rightarrow$scalar}, \texttt{statement$\rightarrow$statements}, \texttt{unconditional$\rightarrow$if}, \texttt{scalar$\rightarrow$array}, \texttt{if$\rightarrow$while}, and \texttt{expression$\rightarrow$function}. 
The LLM is also explicitly instructed to avoid exact input matching, memorized lookup tables, and other hard-coded solutions tied to seen examples. This makes the mutation operator prefer the smallest hypothesis that explains the current evidence before moving to more complex control flow.

We view this design as an evolutionary realization of the classical Red-Green-Refactor lifecycle. When the parent still fails many visible examples, the TPP prior encourages minimal edits that establish a basic working hypothesis. As more examples become satisfied, the LLM is allowed to introduce branches, collections, or loops only when simpler transformations are insufficient. Once the current visible slice is already solved, the search pressure shifts toward removing unnecessary structure rather than adding more case-specific logic. In this way, TDD-style minimal generalization is not run outside evolution; it becomes the mutation prior inside the DIO-Agent loop itself.

\subsection{Error-Grounded Feedback}

Since many candidate programs can obtain similar accuracy scores while failing for very different reasons, DIO-Agent  turns execution results into structured feedback artifacts that are passed back into subsequent mutation prompts.

For a candidate program $p$ at curriculum stage $s$, the evaluator executes $p$ on two sets: the currently visible curriculum set $\mathcal{C}_s$, and a replay set $\mathcal{R}_s$ sampled from examples introduced in earlier stages. The evaluator returns not only accuracies, but also a small number of concrete failure cases from both sets, including mismatched outputs or runtime errors. These artifacts are attached to the evaluated program and, in later iterations, are injected into the prompt of its descendants together with an inherited feedback chain from ancestor programs. As a result, the LLM receives explicit debugging evidence rather than only learning from a scalar reward.

The stage-level selection score is defined as:
\begin{equation}
\mathcal{S}_s(p)=\mathrm{Acc}_{\mathrm{curr}}^{(s)}(p)-\lambda_c\,\Omega_{\mathrm{comp}}(p)-\lambda_h\,\Omega_{\mathrm{hard}}(p),
\end{equation}
where $\mathrm{Acc}_{\mathrm{curr}}^{(s)}$ is the accuracy on the current curriculum set, $\Omega_{\mathrm{comp}}$ penalizes complex programs (e.g., Token Length or Time complexity), and $\Omega_{\mathrm{hard}}$ is a penalty of memorization (e.g., overlap between the provided I/Os and the generated code.
We set $\lambda_c=\lambda_h=0.1$ by default. 

\section{Experiments}
\label{sec:experiments}

We conduct extensive experiments on IO2CodeBench spanning difficulty levels, comparing DIO-Agent against both traditional Programming-by-Example (PBE) methods and mainstream evolution-agent baselines. We also provide the generalizability across different base models, test-time scaling strategy comparison, ablation studies, and further analysis. 
More comprehensive experiments and detailed setups are provided in Appendix~\ref{sec:stage-wise-analysis}-\ref{sec: setup}.

\begin{table}[t]
\centering
\caption{Main results between DIO-Agent and seven baselines on the IO2CodeBench, where the best results are shown in \textbf{bold}.
}
\label{tab:main_results}
\resizebox{\textwidth}{!}{
\begin{tabular}{@{}lcccccc@{}}
\toprule & \multicolumn{6}{c}{IO2CodeBench}\\
\cmidrule(l){2-7}
Method & Base & Algorithm  & Geometry & Multimodal & Average & Token/iter \\
\midrule
Direct & 35.48 & 48.57 & 33.33 & 11.33 & 32.18 & -\\
\midrule
PBE~\citep{gulwani2016programming} & 20.97 & 20.00 & 0 & 17.33 & 14.58 & - \\
E-PBE~\citep{verbruggen2025execution} & 51.61 & 60.00 & 22.22 & 20.67 & 38.63 & 1915.14\\
TDD-guided Generation & 20.97 & 48.57 & 61.11 & 19.33 & 37.50 & \textbf{544.90}\\
\midrule
FunSearch~\citep{romera2024mathematical} & 46.77 & 65.71 & 38.89 & 28.67 & 45.01 & 2911.99 \\
AlphaEvolve~\citep{novikov2025alphaevolve} & 51.61 & 60.00 & 55.56 & 22.00 & 47.29 & 4130.42 \\
CodeEvolve~\citep{assumpccao2025codeevolve} & 61.29 & 60.00 & 44.44 & 32.67 & 49.60 & 5829.32 \\
\midrule
\textbf{DIO-Agent} & \textbf{61.29} & \textbf{71.43} & \textbf{61.11} & \textbf{40.67} & \textbf{58.63} & 3738.31 \\
\bottomrule
\end{tabular}}
\end{table}

\subsection{Main Results}
The main results on the IO2CodeBench are presented in Table \ref{tab:main_results}. DIO-Agent achieves state-of-the-art performance across all difficulty levels, consistently outperforming both traditional PBE methods and modern evolution-agent baselines. Compared to the strongest evolutionary baseline, CodeEvolve, DIO-Agent shows superior generalization, particularly in algorithmic reasoning and geometric computation. In the Base level, DIO-Agent matches the best-performing baseline at 61.29\%, demonstrating its robustness in synthesizing standard patterns. The performance gap widens significantly in harder levels. In the Algorithm level, DIO-Agent reaches 71.43\%, surpassing all baselines. In the Geometry level, DIO-Agent achieves 61.11\%, a substantial improvement over CodeEvolve's 44.44\%. Notably, in the Multimodal level, DIO-Agent attains 40.67\%, nearly doubling AlphaEvolve (22.00\%), suggesting strong capability in managing the noise inherent in multimodal information extraction. Despite its superior performance, DIO-Agent maintains competitive token efficiency. Among high-performing evolutionary methods, DIO-Agent (3738.31 tokens/iter) is considerably more efficient than CodeEvolve (5829.32) and AlphaEvolve (4130.42), indicating that our method relies on a targeted and structured search process rather than brute-force generation.

\subsection{Performance Across Different LLMs}
To evaluate the generalizability of our method, we conduct experiments across three representative LLMs: DeepSeek V3.2, Qwen-3.6-Plus, and Claude-Sonnet-4.6. The results in Table \ref{tab:diff_LLM_results} demonstrate that DIO-Agent consistently enhances program discovery capabilities across all tested models. For instance, using DeepSeek V3.2, DIO-Agent achieves 71.43\% on the Algorithm level, a substantial jump from 17.00\% via Direct prompting and 60.00\% of AlphaEvolve. For Qwen-3.6-Plus, DIO-Agent leads AlphaEvolve by over 14\% in the Algorithm level while maintaining a lower token cost per iteration (7068.3 vs 9274.4). Cross-model comparisons further highlight complementary strengths. 
These results confirm that DIO-Agent provides a model-agnostic performance boost, pushing the ceiling of automated program discovery across diverse LLM backbones.

\begin{table}[t]
\centering
\caption{Performance Across Different LLMs.}
\label{tab:diff_LLM_results}
\resizebox{\textwidth}{!}{
\begin{tabular}{llcccccc}
\toprule
\multirow{2}{*}{Base LLM} & \multirow{2}{*}{Method} & \multicolumn{6}{c}{IO2CodeBench}\\
\cmidrule(l){3-8} & & \multicolumn{1}{l}{Base} & \multicolumn{1}{l}{Algorithm} & \multicolumn{1}{l}{Geometry} & Multimodal & Average & \multicolumn{1}{l}{Token/iter} \\
\midrule
\multirow{3}{*}{deepseek v3.2}     & Direct      & 35.48                    & 48.57                     & 33.33                        &      11.33     &     32.18    & -                      \\ & AlphaEvolve & 51.61                    & 60.00                     & 55.56                        &      22.00      &     47.29    & 4130.42                      \\ & DIO-Agent  & \textbf{61.29} & \textbf{71.43} & \textbf{61.11} & \textbf{40.67} & \textbf{58.63} & \textbf{3738.31}                      \\
\midrule
\multirow{3}{*}{qwen-3.6-plus}     & Direct      & 43.55    & 40.00     & 27.77        &      11.33       &    30.66    & -        \\ & AlphaEvolve & 64.52                    & 57.14                     & 44.44                        &      21.33      &     46.86    & 9274.40                       \\ & DIO-Agent  & \textbf{66.13}                    & \textbf{71.43}                     & \textbf{55.56}                        &      \textbf{28.67}      &    \textbf{55.45}    & \textbf{7068.28}                       \\
\midrule
\multirow{3}{*}{claude-sonnet-4.6} & Direct      & 37.10    & 57.14     & 38.89        &      12.67      &     44.38    & -       \\ & AlphaEvolve & 54.84                    & 68.57                     & 44.44                        &      20.67      &     47.13    & \textbf{2502.24}                       \\ & DIO-Agent  & \textbf{61.29}                    & \textbf{65.71}                     & \textbf{72.22}                        &      \textbf{30.00}      &    \textbf{57.31}    & 2785.67       \\  
\bottomrule
\end{tabular}}
\end{table}

\begin{wraptable}{r}{0.4\textwidth}
\vspace{-10pt}
\centering
\caption{Ablation Study, where each module is removed separately.}
\label{tab:ablation}
\small
\begin{tabular}{lcc}
\toprule
Method & Performance & Token/iter \\
\midrule
DIO-Agent & \textbf{58.63} & 3738.31 \\
\midrule
w/o EF    & 54.61          & 3837.61 \\
w/o TPP   & 53.80          & \textbf{3369.48} \\
w/o CE    & 51.33          & 4244.31 \\
\bottomrule
\end{tabular}
\vspace{-10pt}
\end{wraptable}

\subsection{Ablation Study}
To understand the individual contribution of each component, we conduct an ablation study and report the average performance across all difficulty levels in Table \ref{tab:ablation} (comprehensive results can be found in Appendix \ref{Detailed Ablation Results}).
The most critical component is Curriculum-wise Evolution (CE). Removing it results in the largest performance drop, with the average performance falling from 58.63\% to 51.33\%. Furthermore, its absence increases the token cost per iteration to 4244.31, the highest among all variants, suggesting that the curriculum not only improves accuracy but also makes the search more efficient by pruning unproductive paths early. TPP also plays a vital role, as removing it leads to a noticeable decrease in average performance from 58.63\% to 53.80\%.
Error-Grounded Feedback (EF) serves as a refinement tool. Without it, the average performance drops to 54.61\%, confirming its contribution to the overall search process.

\begin{figure}[h]
    \centering
    \includegraphics[width=0.9\linewidth]{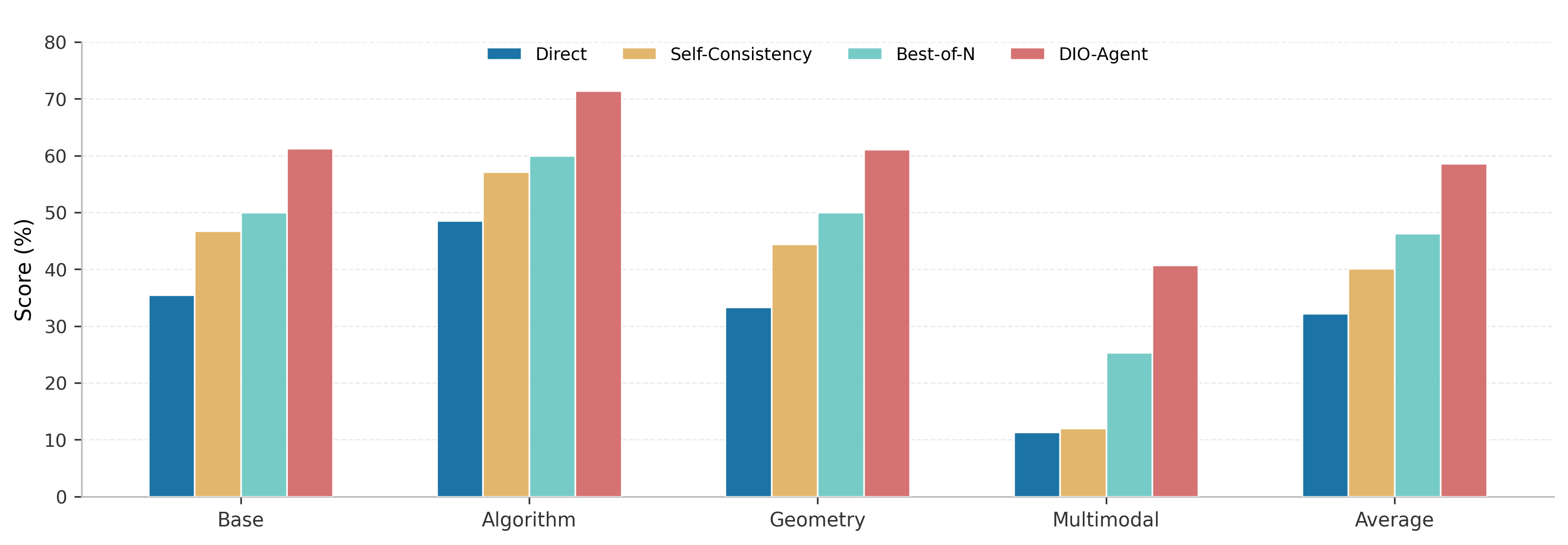}
    \caption{
    Comparison of DIO-Agent with test-time scaling methods.
    }
\label{tab:tts}
\end{figure}

\subsection{Comparison with Test-Time Scaling Methods.}
To demonstrate that the superiority of DIO-Agent stems from its iterative discovery logic rather than a mere increase in sampling budget, we compare it against common test-time scaling baselines: Direct, Best-of-N, and Self-Consistency. For a fair comparison, the sampling budget for these baselines is set to N=40, matching the cumulative candidate generation effort of our 40-iteration search. The results in Figure \ref{tab:tts} reveal a massive performance gap between parallel sampling and iterative search. While Best-of-N is the strongest scaling baseline with an average pass rate of 40.28\%, it still falls nearly 24 percentage points short of DIO-Agent (63.66\%), suggesting that complex problems require systematic refinement rather than repeated independent sampling.
Furthermore, Direct and Self-Consistency perform significantly worse, with average pass rates of only 14.56\% and 25.53\%, respectively. This highlights that for program discovery tasks, the correct rule is often not the most frequent one generated by the model's prior distribution. Thus, simply increasing independent attempts is an inefficient strategy compared to the structured, feedback-driven search of DIO-Agent.

\begin{figure}[h]
    \centering
    \begin{minipage}[t]{0.31\linewidth}
        \centering
        \includegraphics[width=\linewidth]{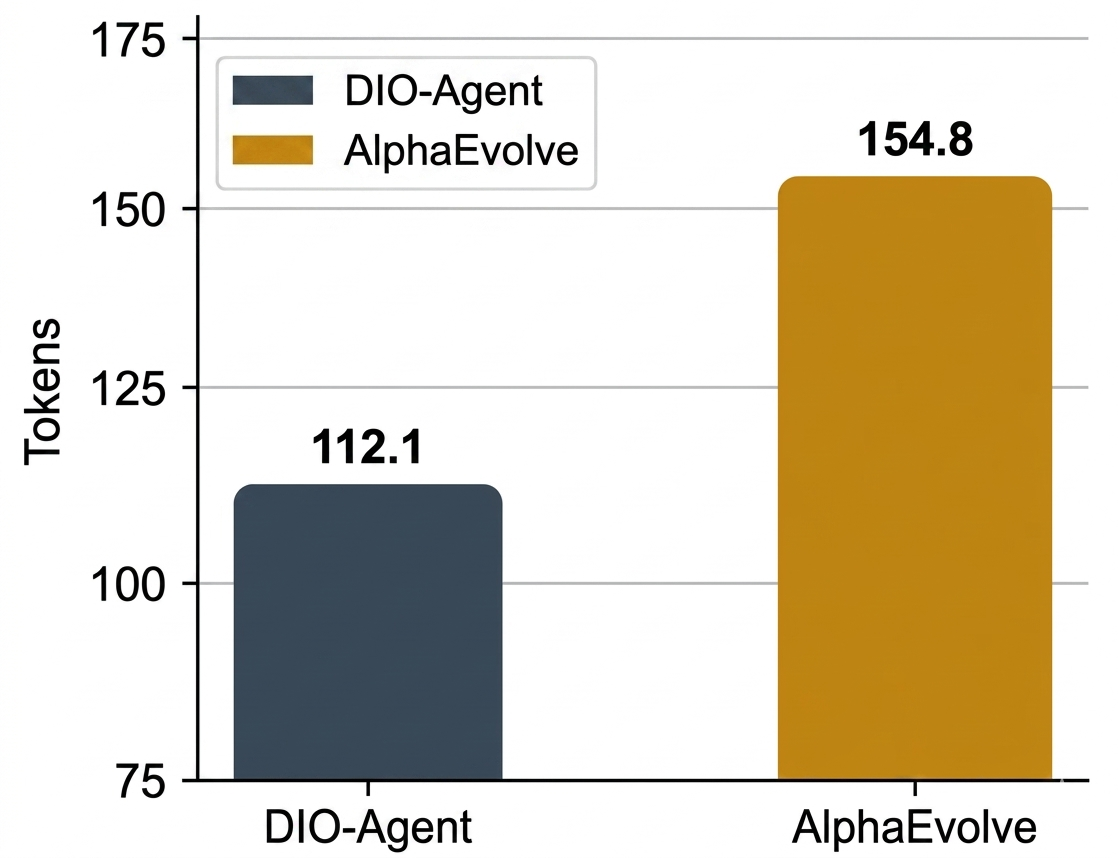}
        \vspace{2pt}
        {\footnotesize (a) Correct Code Length }
    \end{minipage}
    \hfill
    \begin{minipage}[t]{0.31\linewidth}
        \centering
        \includegraphics[width=\linewidth]{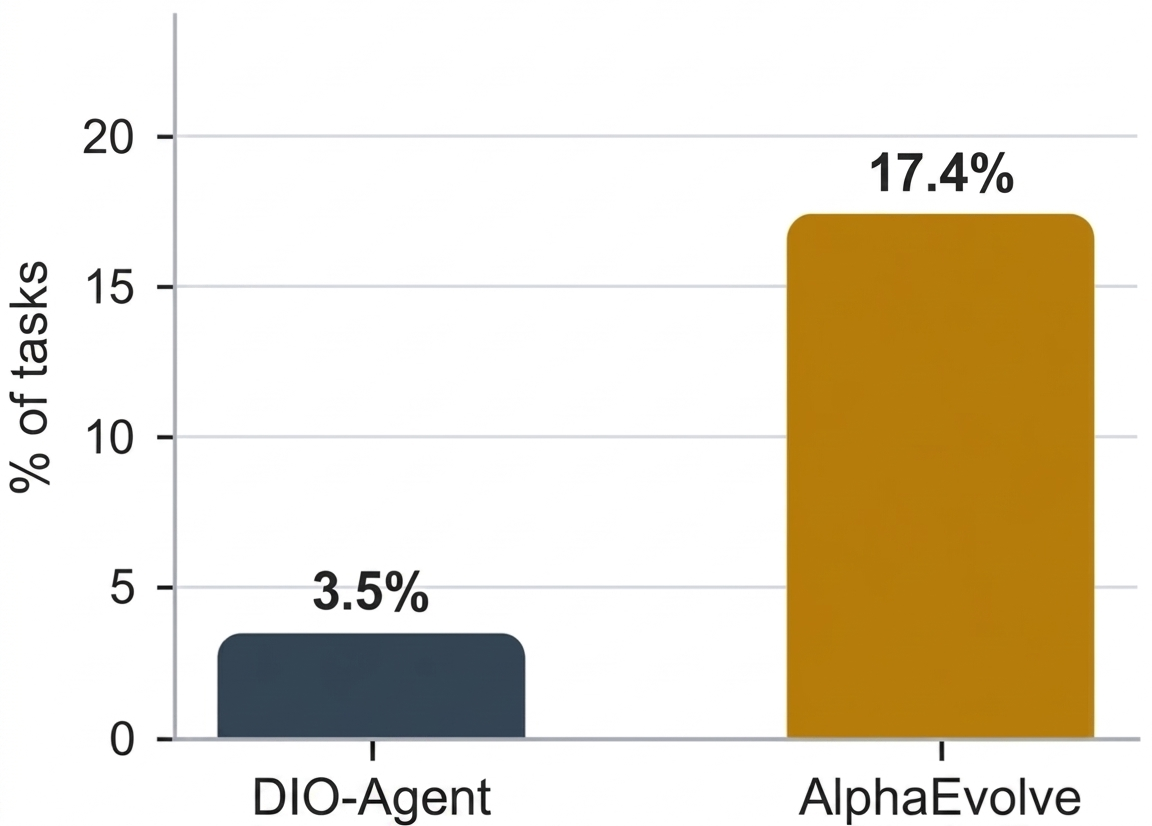}
        \vspace{2pt}
        {\footnotesize (b) Incorrect Early Stop}
    \end{minipage}
    \hfill
    \begin{minipage}[t]{0.36\linewidth}
        \centering
        \includegraphics[width=\linewidth]{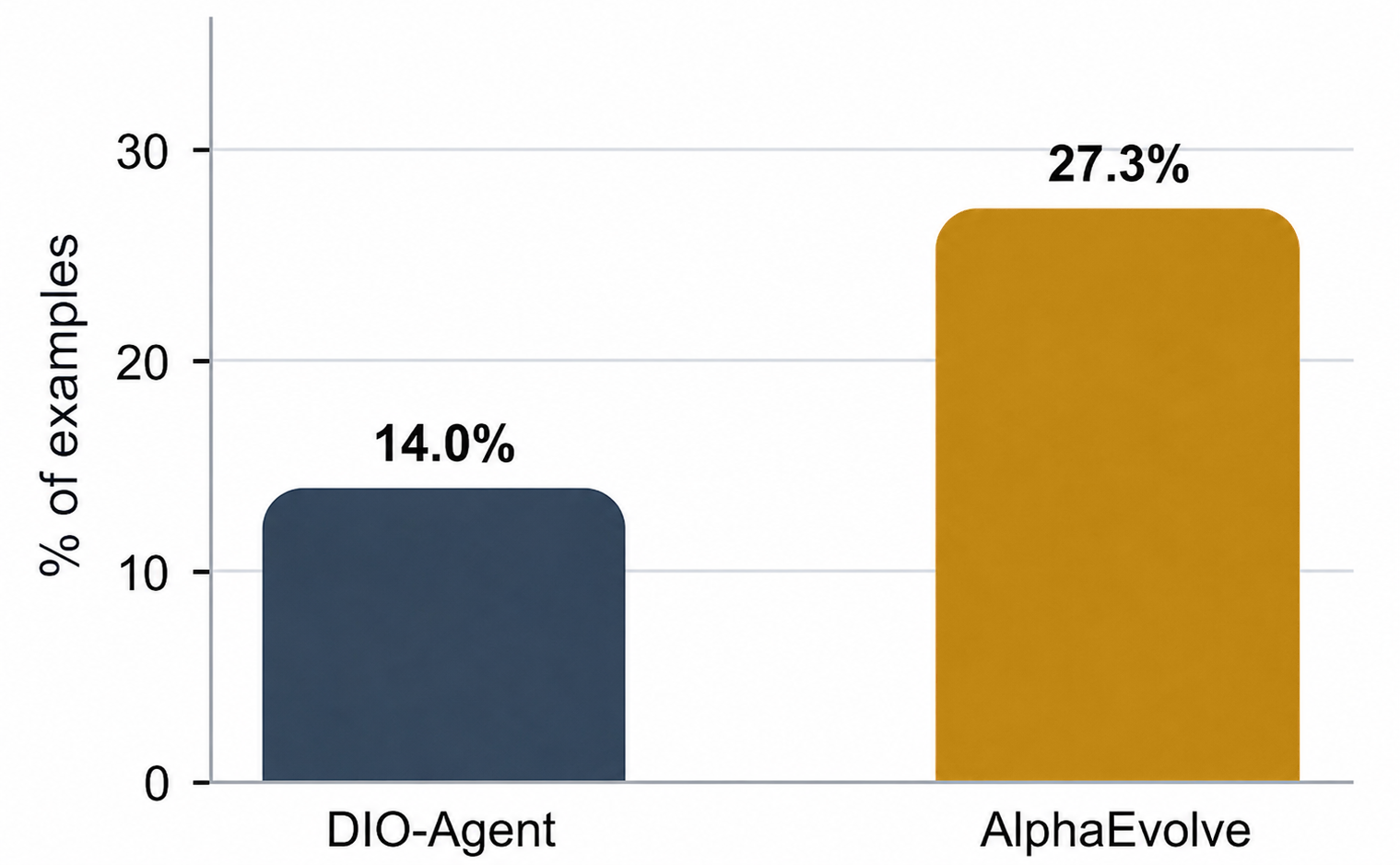}
        \vspace{2pt}
        {\footnotesize (c) Copy Frequency}
    \end{minipage}
    \caption{
    Comparison between DIO-Agent and AlphaEvolve.
    (a) DIO-Agent produces shorter correct programs.
    (b) DIO-Agent reduces incorrect early stopping during search.
    (c) DIO-Agent copies fewer prompt example cases into the generated programs.
    }
\label{fig:mechanism_analysis}
\end{figure}

\subsection{Further Analysis}
To better understand the underlying mechanisms of DIO-Agent, we analyze the evolutionary trajectories and behavioral patterns of the synthesized programs across multiple stages. We provide detailed numerical evidence for these analyses in Appendix~\ref{app:behavioral_analysis}.

\noindent \textbf{Evolutionary Trajectories and Program Complexity.} We observe that program complexity trends are strong indicators of success. Specifically, programs that maintain a stable structure exhibit a success rate of 83.8\%, typically representing simple tasks where the rule is identified early. In contrast, trajectories characterized by continuous simplification achieve a high success rate of 94.4\%; after normalizing for task difficulty, this simplification signal emerges as a robust indicator of genuine discovery. While a complex-then-simple trajectory also shows a positive signal (66.7\% success rate), representing an exploratory path for difficult tasks, the poorest outcomes are associated with monotonic complexity growth. The latter typically reflects patch-work programming, where the model overfits visible examples by stacking logic rather than finding a generalizable rule. Detailed trajectory statistics and the difficulty-adjusted analysis are provided in Table~\ref{tab:dio_complexity_trajectory_success} and Table~\ref{tab:dio_difficulty_adjusted_trajectory}.

\noindent \textbf{Generalization vs. Overfitting.} Achieving a perfect score on visible examples does not necessarily imply the discovery of the underlying logic. Our data shows that 42 runs failed the held-out test cases; among them,
36 achieved a 1.0 accuracy on the final curriculum stage. These failed programs were significantly more verbose, with an average length of 833.9 characters, compared to the 249.8 characters of successful runs. This phenomenon is particularly prevalent in Arithmetic tasks, where models often produce increasingly long, convoluted programs that achieve perfect training scores but collapse upon evaluation against held-out data. Genuine discovery is thus characterized by brevity and cross-stage stability rather than mere performance on visible examples. We provide the corresponding visible-stage overfitting statistics and task-family breakdown in Tables~\ref{tab:dio_visible_overfit} and~\ref{tab:dio_overfit_by_family}.

\noindent \textbf{Statistic Metric.}
We jointly analyze program compactness, premature stopping, and copy behavior in Figure~\ref{fig:mechanism_analysis}. Specifically, 1) As shown in Figure~\ref{fig:mechanism_analysis}(a), DIO-Agent produces substantially shorter correct programs than AlphaEvolve (112.1 vs. 154.8 tokens), indicating reduced accumulation of redundant logic and more concise rule discovery. 
2) From a search perspective, Figure~\ref{fig:mechanism_analysis}(b) shows that only 3.5\% of DIO-Agent tasks terminate early while still remaining unsolved, compared to 17.4\% for AlphaEvolve. This suggests that DIO-Agent is less prone to premature convergence and more likely to continue exploration until a valid rule is discovered. 3) Figure~\ref{fig:mechanism_analysis}(c) measures the proportion of prompt example cases that are directly copied into the generated program. DIO-Agent exhibits a substantially lower copy ratio than AlphaEvolve (14.0\% vs. 27.3\%), indicating that it is less prone to memorizing visible examples through explicit case replication and instead more effectively captures the underlying transformation rules. In summary, these results suggest that DIO-Agent promotes concise and generalizable program discovery while reducing both premature stopping and case-specific memorization.

\subsection{Autonomous Discovery via Active Exploration}
In this section, we transition from a passive evaluation setting to an Active Exploration paradigm. In this configuration, the model is no longer provided with fixed input-output pairs. Instead, it is given access to an encapsulated Ground Truth oracle. The model must autonomously design its own input cases, query the oracle for the corresponding outputs, and construct a curriculum of I/O examples to identify the underlying patterns.
To evaluate the feasibility of this setting, we selected a representative subset comprising 25\% of tasks from each level. This autonomous discovery mode significantly extends the search horizon for each task, as the model must not only synthesize code but also strategically select informative inputs that resolve logical ambiguities.

\begin{table}[h]
    \centering
    \caption{Performance comparison under Autonomous Discovery setting}
    \begin{tabular}{lccccc}
    \toprule
    Methods & Base & Algorithm & Geometry & Multimodal & Average \\
    \midrule
    AlphaEvolve & 38.67 & 64.44  & 60.00 & 23.33 & 46.61\\
    \textbf{DIO-Agent} & \textbf{52.00} &\textbf{ 76.67} & \textbf{64.67} & \textbf{50.00} & \textbf{60.84} \\
    \bottomrule
    \end{tabular}

    \label{tab: autonomous}
\end{table}

The results in Table \ref{tab: autonomous} demonstrate that our DIO-Agent significantly outperforms the AlphaEvolve baseline in this interactive setting, achieving an average pass rate of 60.84\% compared to 46.61\%. This suggests that when given the freedom to construct its own testing environment, the agent can more effectively probe complex algorithmic boundaries than when it is restricted to a static set of provided examples.
Furthermore, DIO-Agent shows remarkable robustness in the Multimodal level. This result indicates that active exploration allows the agent to systematically interrogate the multimodal LLM tool, refining its understanding of image-to-numerical mappings through targeted trial and error. Overall, the superior performance across all levels validates that autonomous input generation acts as a powerful catalyst for program discovery, enabling the agent to resolve ambiguities and verify hypotheses more rigorously than passive learning methods.

\section{Related Work}
\label{sec:related}

\textbf{Programming by Example (PBE).} Programming by Example aims to synthesize programs from input-output examples. Classical PBE systems typically define a domain-specific language, search for programs that satisfy the provided examples, and rank candidate programs to resolve ambiguity~\citep{gulwani2016programming}. Representative systems include FlashFill for spreadsheet string transformations~\citep{gulwani2011automating}, Microsoft PROSE for building DSL-based PBE tools~\citep{polozov2015flashmeta}, and later frameworks that reduce the engineering cost of constructing domain-specific synthesizers~\citep{wu2023programming}. With the rise of large language models, recent work revisits PBE by using LLMs as program generators, search policies, or repair agents. These studies show that LLMs can improve PBE performance when combined with synthetic training data~\citep{li2024programming},table-transformation prompting and hybrid solving~\citep{zhang2025pbe}, execution-guided within-prompt search~\citep{verbruggen2025execution}, compositional decomposition~\citep{khan2025llm}, and iterative example-based evaluation~\citep{fu2025first}. However, existing LLM-PBE work is still mostly centered on traditional PBE domains such as strings, lists, tables, and symbolic rewrite rules, leaving algorithmic reasoning, spatial reasoning, and higher-level executable rule discovery relatively underexplored.

\textbf{Evolutionary Algorithms and LLM-guided Program Evolution.} Evolutionary algorithms search over candidate solutions through selection, mutation, and recombination, and have developed from classical genetic algorithms to genetic programming and neuroevolution~\citep{sampson1976adaptation,koza1994genetic,stanley2002evolving}. Recent systems combine this search paradigm with large language models by treating LLMs as semantic mutation operators over code. \textit{FunSearch} first demonstrated that LLM-generated programs can be evolved with automatic evaluators to discover new mathematical constructions~\citep{romera2024mathematical}, while \textit{AlphaEvolve} further generalized this idea into an evolutionary coding agent that edits seed programs, executes candidates, scores them with user-defined evaluators, and selects promising variants for further improvement~\citep{novikov2025alphaevolve}. Follow-up work has mainly developed in two directions: applying AlphaEvolve-style loops to new domains such as theoretical computer science, compiler optimization, and multi-agent learning~\citep{nagda2025reinforced,chen2026magellan,li2026discovering}\nocite{hsu2026opencollab,liu2026sprout}, or improving the evolutionary procedure itself through island-based search, modular improvement and perturbation, or meta-evolution~\citep{assumpccao2025codeevolve,kravatskiy2026improvevolve,liu2026evox}. However, these works typically start from a known seed program and evolve it toward a predefined objective. Inferring unknown behavioral rules directly from raw observations remains largely unexplored.

\section{Conclusion}
\label{sec:conclusion}

In this paper, we argued that IO2Code is an important but unexplored task different from the mainstream NL2Code task, where NL2Code evaluates the composition and reuse ability of LLMs, while IO2Code evaluates the induction and discovery ability. To this end, we established an IO2CodeBench spanning multiple difficulty levels and proposed DIO-Agent, a discovery agent that tackles IO2Code through structured evolution guided by Transformation Priority Premise. Extensive experiments show that DIO-Agent consistently outperforms existing program-by-example and evolution-agent baselines across all difficulty levels and various LLMs, while substantially surpassing test-time scaling strategies with equivalent sampling budgets. We hope that our work advances agentic reverse engineering and inspires further exploration of program discovery beyond natural-language-driven code generation.

\bibliographystyle{preprint}
\bibliography{references}

\newpage
\appendix

\section{Stage-wise Analysis of Curriculum Refinement}
\label{sec:stage-wise-analysis}

We distinguish between the strict task-level success metric used for final evaluation and a diagnostic sample-level pass ratio used only for analyzing the search process. In the final evaluation, we follow the strict all-pass criterion: a task is considered solved only if the generated program passes all held-out tests for that task. If any held-out test case fails, the task is counted as unsolved. We conduct a stratified random sampling of over 30 non-multimodal tasks according to task categories. Under the strict all-pass criterion, DIO-Agent solves 15 out of 30 tasks, while AlphaEvolve solves 11 out of 30 tasks.

To better understand how the curriculum affects the search process before a task is fully solved, we additionally report a diagnostic metric: the sample-level pass ratio under the full evaluator. This metric measures the fraction of held-out test cases passed by the current best program for each task. It is not used to determine whether a task is solved; instead, it allows us to observe partial progress when a program has not yet passed all held-out tests.

\begin{figure}[h]
    \centering
    \includegraphics[width=0.82\linewidth]{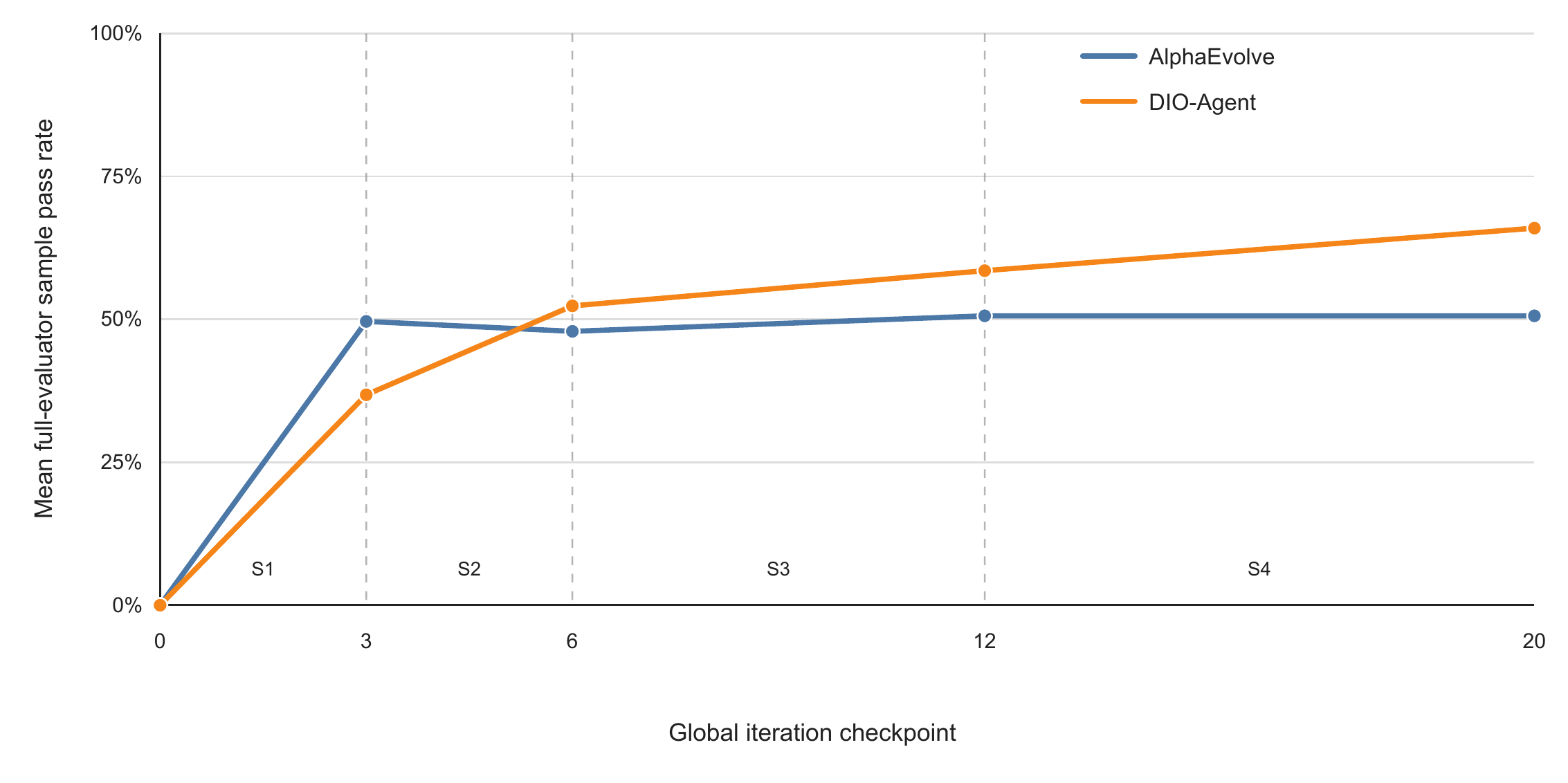}
    \caption{
    Mean full-evaluator sample pass rate across global iteration checkpoints.
    DIO-Agent starts below AlphaEvolve at the first checkpoint, but its performance improves steadily as the curriculum progresses. In contrast, AlphaEvolve remains nearly flat after the early checkpoint.
    }
    \label{fig:stage-pass-ratio}
\end{figure}

As shown in Figure~\ref{fig:stage-pass-ratio}, DIO-Agent exhibits a clear upward trajectory across curriculum stages. Its mean full-evaluator sample pass ratio increases from 36.8\% after stage 1 to 52.3\% after stage 2, 58.5\% after stage 3, and finally 65.9\% after stage 4. In contrast, AlphaEvolve remains around 50\% under the same checkpoints, with sample pass ratios of 49.6\%, 47.9\%, 50.6\%, and 50.6\%, respectively. Therefore, although DIO-Agent is initially lower than AlphaEvolve at the first checkpoint, it continues to benefit from later curriculum stages and eventually surpasses AlphaEvolve.

We further analyze how each checkpoint changes the per-task sample-level pass ratio. A task is counted as improved if its sample pass ratio increases compared with the previous checkpoint, and counted as regressed if its sample pass ratio decreases. This analysis helps separate two effects: whether the overall average performance improves, and how many individual tasks receive additional gains at each stage.

\begin{figure}[h]
    \centering
    \begin{subfigure}[t]{0.48\linewidth}
        \centering
        \includegraphics[width=\linewidth]{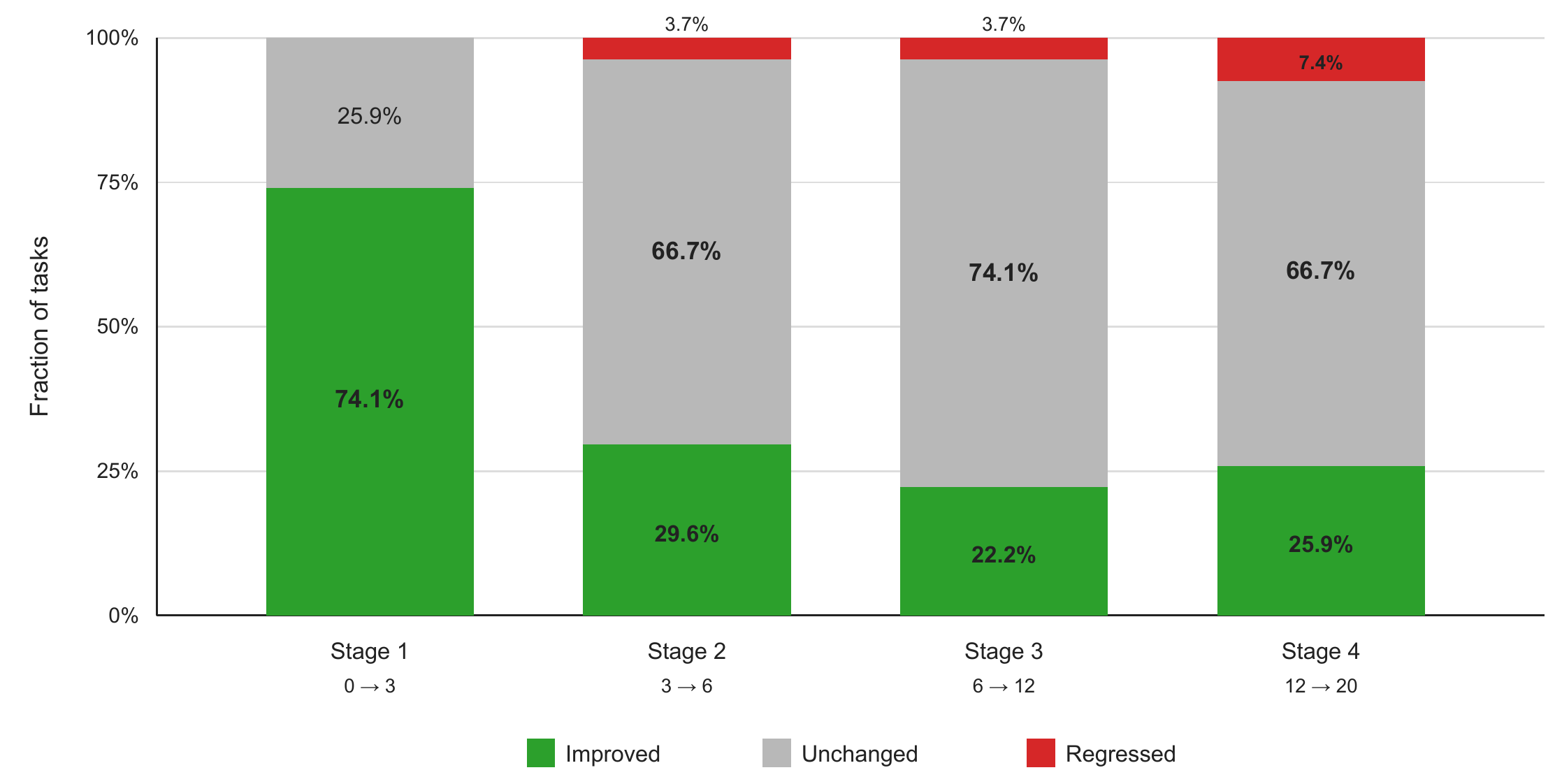}
        \caption{DIO-Agent}
        \label{fig:dio-stage-change}
    \end{subfigure}
    \hfill
    \begin{subfigure}[t]{0.48\linewidth}
        \centering
        \includegraphics[width=\linewidth]{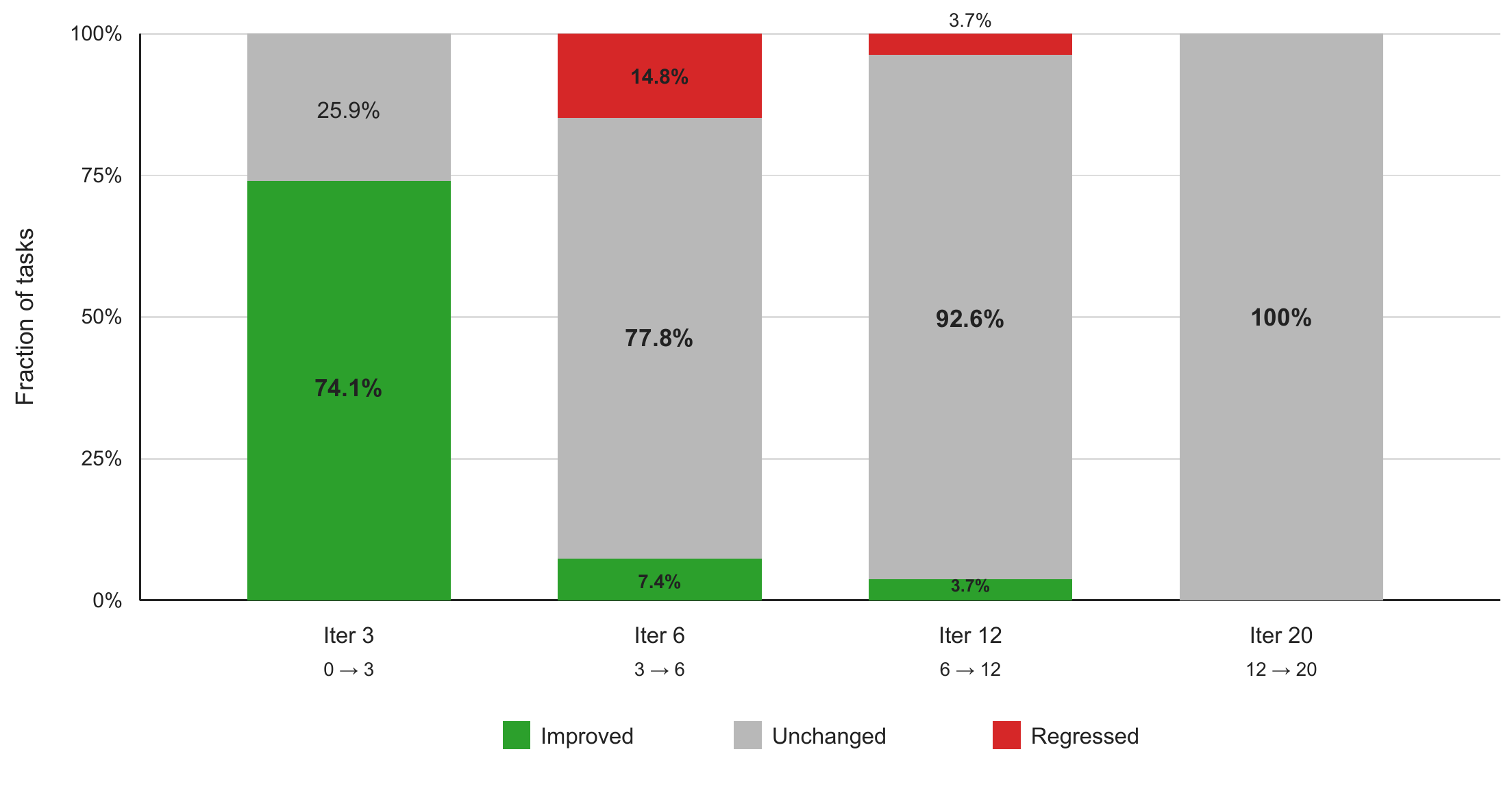}
        \caption{AlphaEvolve}
        \label{fig:alpha-stage-change}
    \end{subfigure}
    \caption{
    Per-checkpoint task-level changes measured by the diagnostic sample-level pass ratio.
    Green indicates tasks whose sample pass ratio improves compared with the previous checkpoint, gray indicates unchanged tasks, and red indicates regressed tasks.
    }
    \label{fig:stage-change-comparison}
\end{figure}

For DIO-Agent, stage 1 improves 22 out of 30 tasks and does not regress any task, suggesting that the early curriculum helps the agent quickly form useful initial hypotheses from simple examples. In later stages, the number of additionally improved tasks becomes smaller: 8, 6, and 7 tasks improve after stages 2, 3, and 4, respectively. Meanwhile, only a few tasks regress, with 1, 1, and 2 regressions in stages 2--4. This does not mean that DIO-Agent becomes worse in later stages, since its mean sample pass ratio continues to increase. Instead, it indicates that the marginal gain of later stages becomes smaller. The role of the curriculum gradually shifts from making many tasks immediately better to validating and refining the hypotheses discovered in earlier stages using more diverse and more difficult examples.

AlphaEvolve shows a different pattern. Its improvement is concentrated at the early checkpoint: after iteration 3, it improves 21 out of 30 tasks. However, its later improvements quickly stagnate. After iterations 6, 12, and 20, only 2, 1, and 0 tasks continue to improve, respectively, and iteration 6 even causes 4 tasks to regress. This contrast suggests that DIO-Agent's curriculum stages do more than simply increase the number of search iterations. By progressively expanding the visible examples, the curriculum provides continued validation pressure, which helps the search process filter out unstable programs and refine promising hypotheses.

\section{Comprehensive results}

\begingroup
\scriptsize
\setlength{\tabcolsep}{2.5pt}
\renewcommand{\arraystretch}{0.88}
\begin{longtable}{lllcccccccc}
\caption{Comprehensive results of per-task pass ratios on training and test examples, completed with additional baselines.}\label{tab:overall-statistic-full-compact} \\
\toprule
Category & Task & Set & Direct & PBE & E-PBE & TDD & FunSearch & AlphaEvolve & CodeEvolve & DIO-Agent \\
\midrule
\endfirsthead
\toprule
Category & Task & Set & Direct & PBE & E-PBE & TDD & FunSearch & AlphaEvolve & CodeEvolve & DIO-Agent \\
\midrule
\endhead
\midrule
\multicolumn{11}{r}{Continued on next page} \\
\endfoot
\bottomrule
\endlastfoot
Base & Abs\_Current & Train & 1.00 & 1.00 & 1.00 & 1.00 & 1.00 & 1.00 & 1.00 & 1.00 \\
 &  & Test & 1.00 & 1.00 & 1.00 & 1.00 & 1.00 & 1.00 & 1.00 & 1.00 \\
 & Abs\_Diff & Train & 0.00 & 0.00 & 1.00 & 1.00 & 1.00 & 1.00 & 1.00 & 1.00 \\
 &  & Test & 0.00 & 0.00 & 1.00 & 0.87 & 0.93 & 1.00 & 1.00 & 1.00 \\
 & Add\_Mod\_3 & Train & 0.00 & 0.00 & 0.00 & 1.00 & 0.00 & 0.00 & 1.00 & 1.00 \\
 &  & Test & 0.00 & 0.00 & 0.00 & 0.87 & 0.00 & 0.00 & 0.93 & 1.00 \\
 & Add\_Mod\_4 & Train & 0.00 & 0.00 & 1.00 & 1.00 & 1.00 & 1.00 & 1.00 & 1.00 \\
 &  & Test & 0.00 & 0.00 & 0.00 & 0.87 & 0.00 & 0.00 & 0.93 & 1.00 \\
 & Add\_Mod\_5 & Train & 0.00 & 0.00 & 1.00 & 0.25 & 1.00 & 1.00 & 0.25 & 0.25 \\
 &  & Test & 0.00 & 0.00 & 0.00 & 0.00 & 0.00 & 0.00 & 0.00 & 0.00 \\
 & Add\_Mod\_6 & Train & 0.00 & 0.00 & 1.00 & 1.00 & 1.00 & 1.00 & 1.00 & 1.00 \\
 &  & Test & 0.00 & 0.00 & 0.00 & 0.00 & 0.00 & 0.00 & 0.00 & 0.00 \\
 & Add\_Mod\_7 & Train & 1.00 & 1.00 & 1.00 & 1.00 & 1.00 & 1.00 & 1.00 & 1.00 \\
 &  & Test & 0.67 & 0.60 & 0.60 & 0.87 & 0.67 & 0.67 & 1.00 & 1.00 \\
 & Add\_Mod\_8 & Train & 0.75 & 0.75 & 1.00 & 1.00 & 1.00 & 1.00 & 1.00 & 1.00 \\
 &  & Test & 0.27 & 0.20 & 1.00 & 0.87 & 0.93 & 1.00 & 1.00 & 1.00 \\
 & Alternating\_Last3 & Train & 0.38 & 0.38 & 0.38 & 0.75 & 0.38 & 0.38 & 0.75 & 0.75 \\
 &  & Test & 0.27 & 0.20 & 0.06 & 0.00 & 0.13 & 0.13 & 0.00 & 0.07 \\
 & Alternating\_Last4 & Train & 0.25 & 0.25 & 1.00 & 1.00 & 1.00 & 1.00 & 1.00 & 1.00 \\
 &  & Test & 0.07 & 0.00 & 0.60 & 0.34 & 0.67 & 0.67 & 0.40 & 0.47 \\
 & Balanced\_Parenthesis & Train & 1.00 & 1.00 & 1.00 & 1.00 & 1.00 & 1.00 & 1.00 & 1.00 \\
 &  & Test & 0.80 & 0.73 & 0.73 & 0.67 & 0.80 & 0.80 & 1.00 & 0.80 \\
 & Base\_3\_Addition & Train & 0.00 & 0.00 & 0.25 & 1.00 & 0.27 & 0.25 & 1.00 & 1.00 \\
 &  & Test & 0.00 & 0.00 & 0.20 & 0.07 & 0.27 & 0.27 & 0.13 & 0.20 \\
 & Base\_4\_Addition & Train & 0.50 & 0.50 & 0.38 & 1.00 & 0.38 & 0.38 & 1.00 & 1.00 \\
 &  & Test & 0.00 & 0.00 & 0.06 & 0.00 & 0.13 & 0.13 & 0.06 & 0.13 \\
 & Base\_5\_Addition & Train & 0.25 & 0.25 & 0.25 & 0.88 & 0.25 & 0.25 & 0.88 & 0.88 \\
 &  & Test & 0.00 & 0.00 & 0.06 & 0.00 & 0.13 & 0.13 & 0.00 & 0.00 \\
 & Base\_6\_Addition & Train & 0.00 & 0.00 & 0.25 & 1.00 & 0.25 & 0.25 & 1.00 & 1.00 \\
 &  & Test & 0.00 & 0.00 & 0.00 & 0.00 & 0.00 & 0.00 & 0.00 & 0.07 \\
 & Base\_7\_Addition & Train & 0.13 & 0.13 & 0.38 & 1.00 & 0.38 & 0.38 & 1.00 & 1.00 \\
 &  & Test & 0.00 & 0.00 & 0.00 & 0.00 & 0.00 & 0.00 & 0.00 & 0.00 \\
 & Binary\_Addition & Train & 0.50 & 0.50 & 0.50 & 1.00 & 0.50 & 0.50 & 1.00 & 1.00 \\
 &  & Test & 0.20 & 0.13 & 0.20 & 0.60 & 0.27 & 0.27 & 0.66 & 0.73 \\
 & Bit\_Dot\_Prod\_Mod2 & Train & 1.00 & 1.00 & 1.00 & 1.00 & 1.00 & 1.00 & 1.00 & 1.00 \\
 &  & Test & 0.47 & 0.40 & 0.40 & 0.34 & 0.47 & 0.47 & 0.40 & 0.47 \\
 & Bit\_Palindrome & Train & 0.00 & 0.00 & 1.00 & 1.00 & 1.00 & 1.00 & 1.00 & 1.00 \\
 &  & Test & 0.00 & 0.00 & 0.00 & 0.00 & 0.00 & 0.00 & 0.00 & 0.07 \\
 & Bit\_Shift\_Right & Train & 0.63 & 0.63 & 1.00 & 1.00 & 1.00 & 1.00 & 1.00 & 1.00 \\
 &  & Test & 0.27 & 0.20 & 1.00 & 0.87 & 0.93 & 1.00 & 1.00 & 1.00 \\
 & Bitwise\_And & Train & 1.00 & 1.00 & 1.00 & 1.00 & 1.00 & 1.00 & 1.00 & 1.00 \\
 &  & Test & 1.00 & 0.93 & 1.00 & 0.87 & 1.00 & 0.33 & 1.00 & 1.00 \\
 & Bitwise\_Not & Train & 1.00 & 1.00 & 1.00 & 1.00 & 1.00 & 1.00 & 1.00 & 1.00 \\
 &  & Test & 1.00 & 1.00 & 1.00 & 1.00 & 1.00 & 1.00 & 1.00 & 1.00 \\
 & Bitwise\_Or & Train & 1.00 & 1.00 & 1.00 & 1.00 & 1.00 & 1.00 & 1.00 & 1.00 \\
 &  & Test & 1.00 & 1.00 & 1.00 & 1.00 & 1.00 & 1.00 & 1.00 & 1.00 \\
 & Bitwise\_Xor & Train & 0.00 & 0.13 & 1.00 & 1.00 & 1.00 & 1.00 & 1.00 & 1.00 \\
 &  & Test & 0.20 & 0.13 & 1.00 & 0.87 & 1.00 & 1.00 & 1.00 & 1.00 \\
 & Current\_Number & Train & 1.00 & 1.00 & 1.00 & 1.00 & 1.00 & 1.00 & 1.00 & 1.00 \\
 &  & Test & 1.00 & 1.00 & 1.00 & 1.00 & 1.00 & 1.00 & 1.00 & 1.00 \\
 & Diff\_Abs\_Values & Train & 0.00 & 0.00 & 1.00 & 1.00 & 1.00 & 1.00 & 1.00 & 1.00 \\
 &  & Test & 0.00 & 0.00 & 0.93 & 0.87 & 0.93 & 1.00 & 1.00 & 1.00 \\
 & Diff\_Last2 & Train & 1.00 & 1.00 & 1.00 & 1.00 & 1.00 & 1.00 & 1.00 & 1.00 \\
 &  & Test & 1.00 & 1.00 & 1.00 & 1.00 & 1.00 & 1.00 & 1.00 & 1.00 \\
 & Dithering & Train & 0.00 & 0.00 & 0.13 & 0.38 & 0.20 & 0.13 & 0.38 & 0.38 \\
 &  & Test & 0.00 & 0.00 & 0.13 & 0.00 & 0.20 & 0.20 & 0.00 & 0.00 \\
 & Div\_3 & Train & 0.00 & 0.00 & 0.38 & 0.50 & 0.38 & 0.38 & 0.50 & 0.50 \\
 &  & Test & 0.00 & 0.00 & 0.00 & 0.00 & 0.07 & 0.07 & 0.00 & 0.00 \\
 & Div\_5 & Train & 0.00 & 0.00 & 1.00 & 1.00 & 1.00 & 1.00 & 1.00 & 1.00 \\
 &  & Test & 0.00 & 0.00 & 0.13 & 0.00 & 0.20 & 0.20 & 0.00 & 0.00 \\
 & Div\_7 & Train & 0.00 & 0.00 & 1.00 & 0.50 & 1.00 & 1.00 & 0.50 & 0.50 \\
 &  & Test & 0.00 & 0.00 & 0.26 & 0.14 & 0.33 & 0.33 & 0.20 & 0.27 \\
 & Evens\_Counter & Train & 0.00 & 0.00 & 0.25 & 1.00 & 0.25 & 0.25 & 1.00 & 1.00 \\
 &  & Test & 0.00 & 0.00 & 0.00 & 0.00 & 0.00 & 0.00 & 0.00 & 0.00 \\
 & Evens\_Detector & Train & 0.00 & 0.00 & 1.00 & 1.00 & 1.00 & 1.00 & 1.00 & 1.00 \\
 &  & Test & 0.00 & 0.00 & 0.93 & 0.87 & 0.93 & 1.00 & 1.00 & 1.00 \\
 & Majority\_0\_1 & Train & 0.00 & 0.00 & 0.63 & 1.00 & 0.63 & 0.63 & 1.00 & 1.00 \\
 &  & Test & 0.00 & 0.00 & 0.33 & 0.00 & 0.40 & 0.40 & 0.06 & 0.13 \\
 & Majority\_0\_2 & Train & 0.50 & 0.50 & 1.00 & 1.00 & 1.00 & 1.00 & 1.00 & 1.00 \\
 &  & Test & 0.40 & 0.33 & 0.46 & 0.47 & 0.53 & 0.53 & 0.53 & 0.60 \\
 & Majority\_0\_3 & Train & 0.88 & 0.88 & 1.00 & 1.00 & 1.00 & 1.00 & 1.00 & 1.00 \\
 &  & Test & 0.80 & 0.73 & 0.66 & 0.74 & 0.73 & 0.73 & 1.00 & 0.87 \\
 & Max\_Seen & Train & 1.00 & 1.00 & 1.00 & 1.00 & 1.00 & 1.00 & 1.00 & 1.00 \\
 &  & Test & 1.00 & 1.00 & 1.00 & 0.87 & 1.00 & 1.00 & 1.00 & 1.00 \\
 & Min\_Seen & Train & 1.00 & 1.00 & 1.00 & 1.00 & 1.00 & 1.00 & 1.00 & 1.00 \\
 &  & Test & 1.00 & 1.00 & 1.00 & 0.87 & 1.00 & 1.00 & 1.00 & 1.00 \\
 & Newton\_Freebody & Train & 1.00 & 1.00 & 1.00 & 1.00 & 1.00 & 1.00 & 1.00 & 1.00 \\
 &  & Test & 1.00 & 1.00 & 1.00 & 1.00 & 1.00 & 1.00 & 1.00 & 1.00 \\
 & Newton\_Gravity & Train & 1.00 & 1.00 & 1.00 & 1.00 & 1.00 & 1.00 & 1.00 & 1.00 \\
 &  & Test & 1.00 & 1.00 & 1.00 & 1.00 & 1.00 & 1.00 & 1.00 & 1.00 \\
 & Newton\_Magnetic & Train & 1.00 & 1.00 & 1.00 & 1.00 & 1.00 & 1.00 & 1.00 & 1.00 \\
 &  & Test & 1.00 & 1.00 & 1.00 & 1.00 & 1.00 & 1.00 & 1.00 & 1.00 \\
 & Newton\_Spring & Train & 1.00 & 1.00 & 1.00 & 1.00 & 1.00 & 1.00 & 1.00 & 1.00 \\
 &  & Test & 1.00 & 1.00 & 1.00 & 1.00 & 1.00 & 1.00 & 1.00 & 1.00 \\
 & Parity\_All & Train & 0.50 & 0.50 & 1.00 & 1.00 & 1.00 & 1.00 & 1.00 & 1.00 \\
 &  & Test & 0.20 & 0.13 & 1.00 & 0.87 & 1.00 & 1.00 & 1.00 & 1.00 \\
 & Parity\_Bits\_Mod2 & Train & 1.00 & 1.00 & 1.00 & 1.00 & 1.00 & 1.00 & 1.00 & 1.00 \\
 &  & Test & 1.00 & 0.93 & 1.00 & 0.87 & 1.00 & 1.00 & 1.00 & 1.00 \\
 & Parity\_Last2 & Train & 1.00 & 1.00 & 1.00 & 1.00 & 1.00 & 1.00 & 1.00 & 1.00 \\
 &  & Test & 1.00 & 1.00 & 1.00 & 0.87 & 1.00 & 1.00 & 1.00 & 1.00 \\
 & Parity\_Last3 & Train & 0.00 & 0.00 & 0.50 & 1.00 & 0.50 & 0.50 & 1.00 & 1.00 \\
 &  & Test & 0.00 & 0.00 & 0.20 & 0.87 & 0.27 & 0.27 & 1.00 & 1.00 \\
 & Parity\_Last4 & Train & 0.63 & 0.63 & 0.63 & 0.75 & 0.63 & 0.63 & 0.75 & 0.75 \\
 &  & Test & 0.40 & 0.33 & 0.33 & 0.20 & 0.40 & 0.40 & 0.26 & 0.33 \\
 & Parity\_Zeros & Train & 0.13 & 0.13 & 1.00 & 0.25 & 1.00 & 1.00 & 0.25 & 0.25 \\
 &  & Test & 0.20 & 0.13 & 0.00 & 0.00 & 0.00 & 0.00 & 0.00 & 0.00 \\
 & Perfect\_Square\_Detector & Train & 0.88 & 0.88 & 1.00 & 1.00 & 1.00 & 1.00 & 1.00 & 1.00 \\
 &  & Test & 0.20 & 0.13 & 0.13 & 0.87 & 0.20 & 0.20 & 0.93 & 1.00 \\
 & Prev1 & Train & 1.00 & 1.00 & 1.00 & 1.00 & 1.00 & 1.00 & 1.00 & 1.00 \\
 &  & Test & 1.00 & 1.00 & 1.00 & 1.00 & 1.00 & 1.00 & 1.00 & 1.00 \\
 & Prev2 & Train & 1.00 & 1.00 & 1.00 & 1.00 & 1.00 & 1.00 & 1.00 & 1.00 \\
 &  & Test & 1.00 & 0.93 & 1.00 & 1.00 & 1.00 & 1.00 & 1.00 & 1.00 \\
 & Prev3 & Train & 1.00 & 1.00 & 1.00 & 1.00 & 1.00 & 1.00 & 1.00 & 1.00 \\
 &  & Test & 1.00 & 0.93 & 1.00 & 1.00 & 1.00 & 1.00 & 1.00 & 1.00 \\
 & Prev4 & Train & 0.25 & 0.25 & 1.00 & 1.00 & 1.00 & 1.00 & 1.00 & 1.00 \\
 &  & Test & 0.20 & 0.13 & 1.00 & 0.67 & 1.00 & 1.00 & 1.00 & 0.80 \\
 & Prev5 & Train & 1.00 & 1.00 & 1.00 & 1.00 & 1.00 & 1.00 & 1.00 & 1.00 \\
 &  & Test & 1.00 & 0.93 & 1.00 & 1.00 & 1.00 & 1.00 & 1.00 & 1.00 \\
 & Previous\_Equals\_Current & Train & 1.00 & 1.00 & 1.00 & 1.00 & 1.00 & 1.00 & 1.00 & 1.00 \\
 &  & Test & 1.00 & 0.93 & 1.00 & 0.87 & 1.00 & 1.00 & 1.00 & 1.00 \\
 & Sum\_All & Train & 1.00 & 1.00 & 1.00 & 1.00 & 1.00 & 1.00 & 1.00 & 1.00 \\
 &  & Test & 1.00 & 0.93 & 1.00 & 0.87 & 1.00 & 1.00 & 1.00 & 1.00 \\
 & Sum\_Last2 & Train & 1.00 & 1.00 & 1.00 & 1.00 & 1.00 & 1.00 & 1.00 & 1.00 \\
 &  & Test & 1.00 & 0.93 & 1.00 & 0.87 & 1.00 & 1.00 & 1.00 & 1.00 \\
 & Sum\_Last3 & Train & 0.25 & 0.25 & 1.00 & 1.00 & 1.00 & 1.00 & 1.00 & 1.00 \\
 &  & Test & 0.13 & 0.06 & 1.00 & 0.87 & 1.00 & 1.00 & 1.00 & 1.00 \\
 & Sum\_Last4 & Train & 0.25 & 0.25 & 1.00 & 1.00 & 1.00 & 1.00 & 1.00 & 1.00 \\
 &  & Test & 0.27 & 0.20 & 1.00 & 0.87 & 1.00 & 1.00 & 1.00 & 1.00 \\
 & Sum\_Last5 & Train & 0.75 & 0.75 & 1.00 & 1.00 & 1.00 & 1.00 & 1.00 & 1.00 \\
 &  & Test & 0.40 & 0.33 & 1.00 & 0.87 & 1.00 & 1.00 & 1.00 & 1.00 \\
 & Sum\_Last6 & Train & 0.88 & 0.88 & 1.00 & 1.00 & 1.00 & 1.00 & 1.00 & 1.00 \\
 &  & Test & 0.53 & 0.46 & 1.00 & 0.87 & 1.00 & 1.00 & 1.00 & 1.00 \\
 & Sum\_Last7 & Train & 0.00 & 0.00 & 1.00 & 1.00 & 1.00 & 1.00 & 1.00 & 1.00 \\
 &  & Test & 0.00 & 0.00 & 1.00 & 0.87 & 1.00 & 0.60 & 1.00 & 1.00 \\
\midrule
Algorithm & AllPositive & Train & 1.00 & 1.00 & 1.00 & 1.00 & 1.00 & 1.00 & 1.00 & 1.00 \\
 &  & Test & 0.80 & 0.73 & 0.73 & 0.87 & 1.00 & 0.80 & 0.93 & 1.00 \\
 & BalancedParentheses & Train & 1.00 & 1.00 & 1.00 & 1.00 & 1.00 & 1.00 & 1.00 & 1.00 \\
 &  & Test & 0.47 & 0.40 & 0.40 & 0.40 & 0.47 & 0.47 & 0.46 & 0.53 \\
 & BinaryAddition & Train & 1.00 & 1.00 & 1.00 & 1.00 & 1.00 & 1.00 & 1.00 & 1.00 \\
 &  & Test & 1.00 & 0.93 & 1.00 & 1.00 & 1.00 & 1.00 & 1.00 & 1.00 \\
 & ClimbStairs & Train & 1.00 & 1.00 & 1.00 & 1.00 & 1.00 & 1.00 & 1.00 & 1.00 \\
 &  & Test & 0.93 & 0.86 & 1.00 & 0.87 & 1.00 & 1.00 & 1.00 & 1.00 \\
 & CountChars & Train & 1.00 & 1.00 & 1.00 & 1.00 & 1.00 & 1.00 & 1.00 & 1.00 \\
 &  & Test & 1.00 & 1.00 & 1.00 & 1.00 & 1.00 & 1.00 & 1.00 & 1.00 \\
 & CountDigits & Train & 1.00 & 1.00 & 1.00 & 1.00 & 1.00 & 1.00 & 1.00 & 1.00 \\
 &  & Test & 1.00 & 1.00 & 1.00 & 1.00 & 1.00 & 1.00 & 1.00 & 1.00 \\
 & CountZeros & Train & 1.00 & 1.00 & 1.00 & 1.00 & 1.00 & 1.00 & 1.00 & 1.00 \\
 &  & Test & 1.00 & 1.00 & 1.00 & 1.00 & 1.00 & 1.00 & 1.00 & 1.00 \\
 & Dec2Bin & Train & 0.00 & 0.13 & 1.00 & 1.00 & 1.00 & 1.00 & 1.00 & 1.00 \\
 &  & Test & 0.20 & 0.13 & 1.00 & 0.87 & 1.00 & 1.00 & 0.93 & 1.00 \\
 & Dec2Roman & Train & 1.00 & 1.00 & 1.00 & 1.00 & 1.00 & 1.00 & 1.00 & 1.00 \\
 &  & Test & 1.00 & 0.93 & 1.00 & 1.00 & 1.00 & 1.00 & 1.00 & 1.00 \\
 & Factorial & Train & 0.00 & 0.00 & 1.00 & 1.00 & 1.00 & 1.00 & 1.00 & 1.00 \\
 &  & Test & 0.00 & 0.00 & 0.93 & 1.00 & 1.00 & 1.00 & 0.93 & 1.00 \\
 & Fibonacci & Train & 1.00 & 1.00 & 1.00 & 1.00 & 1.00 & 1.00 & 1.00 & 1.00 \\
 &  & Test & 1.00 & 0.93 & 1.00 & 1.00 & 1.00 & 1.00 & 1.00 & 1.00 \\
 & GCD & Train & 1.00 & 1.00 & 1.00 & 1.00 & 1.00 & 1.00 & 1.00 & 1.00 \\
 &  & Test & 1.00 & 1.00 & 1.00 & 1.00 & 1.00 & 1.00 & 1.00 & 1.00 \\
 & HammingWeight & Train & 1.00 & 1.00 & 1.00 & 1.00 & 1.00 & 1.00 & 1.00 & 1.00 \\
 &  & Test & 1.00 & 0.93 & 1.00 & 1.00 & 1.00 & 1.00 & 1.00 & 1.00 \\
 & HouseRobber & Train & 1.00 & 1.00 & 1.00 & 1.00 & 1.00 & 1.00 & 1.00 & 1.00 \\
 &  & Test & 1.00 & 0.93 & 1.00 & 0.87 & 1.00 & 0.87 & 1.00 & 1.00 \\
 & IntegerDivision & Train & 1.00 & 1.00 & 1.00 & 1.00 & 1.00 & 1.00 & 1.00 & 1.00 \\
 &  & Test & 1.00 & 0.93 & 1.00 & 1.00 & 1.00 & 1.00 & 1.00 & 1.00 \\
 & IsPalindrome & Train & 1.00 & 1.00 & 1.00 & 1.00 & 1.00 & 1.00 & 1.00 & 1.00 \\
 &  & Test & 0.47 & 0.40 & 0.40 & 0.87 & 0.47 & 0.47 & 0.93 & 1.00 \\
 & IsSorted & Train & 1.00 & 1.00 & 1.00 & 1.00 & 1.00 & 1.00 & 1.00 & 1.00 \\
 &  & Test & 0.73 & 0.66 & 0.66 & 0.60 & 0.73 & 0.73 & 0.66 & 0.73 \\
 & KthRoot & Train & 1.00 & 1.00 & 1.00 & 1.00 & 1.00 & 1.00 & 1.00 & 1.00 \\
 &  & Test & 0.87 & 0.80 & 0.20 & 0.14 & 0.27 & 0.27 & 0.20 & 0.27 \\
 & KthRootMultidigit & Train & 1.00 & 1.00 & 1.00 & 1.00 & 1.00 & 1.00 & 1.00 & 1.00 \\
 &  & Test & 0.93 & 0.86 & 0.66 & 0.40 & 0.73 & 0.73 & 0.46 & 0.53 \\
 & LeadingZeros & Train & 0.88 & 0.88 & 1.00 & 1.00 & 1.00 & 1.00 & 1.00 & 1.00 \\
 &  & Test & 0.40 & 0.33 & 0.73 & 0.87 & 0.80 & 0.80 & 0.93 & 1.00 \\
 & ListProduct & Train & 1.00 & 1.00 & 1.00 & 1.00 & 1.00 & 1.00 & 1.00 & 1.00 \\
 &  & Test & 0.93 & 0.86 & 1.00 & 1.00 & 1.00 & 1.00 & 1.00 & 1.00 \\
 & ListSum & Train & 1.00 & 1.00 & 1.00 & 1.00 & 1.00 & 1.00 & 1.00 & 1.00 \\
 &  & Test & 1.00 & 1.00 & 1.00 & 1.00 & 1.00 & 1.00 & 1.00 & 1.00 \\
 & LongestIncreasingSubseqLen & Train & 1.00 & 1.00 & 1.00 & 1.00 & 1.00 & 1.00 & 1.00 & 1.00 \\
 &  & Test & 1.00 & 0.93 & 0.46 & 0.40 & 0.53 & 0.53 & 0.46 & 0.53 \\
 & Modulo & Train & 1.00 & 1.00 & 1.00 & 1.00 & 1.00 & 1.00 & 1.00 & 1.00 \\
 &  & Test & 1.00 & 1.00 & 1.00 & 1.00 & 1.00 & 1.00 & 1.00 & 1.00 \\
 & Multiply & Train & 1.00 & 1.00 & 1.00 & 1.00 & 1.00 & 1.00 & 1.00 & 1.00 \\
 &  & Test & 1.00 & 1.00 & 1.00 & 1.00 & 1.00 & 1.00 & 1.00 & 1.00 \\
 & PolishRPN & Train & 0.50 & 0.60 & 0.88 & 0.75 & 0.88 & 0.88 & 0.75 & 0.75 \\
 &  & Test & 0.67 & 0.60 & 0.66 & 0.60 & 0.73 & 0.73 & 0.66 & 0.73 \\
 & PolishRPN2 & Train & 0.50 & 0.66 & 0.66 & 0.63 & 0.73 & 0.50 & 0.63 & 0.63 \\
 &  & Test & 0.73 & 0.66 & 0.66 & 0.07 & 0.73 & 0.73 & 0.13 & 0.20 \\
 & PolishRPN3 & Train & 0.00 & 0.06 & 0.75 & 0.50 & 0.75 & 0.75 & 0.50 & 0.50 \\
 &  & Test & 0.13 & 0.06 & 0.00 & 0.00 & 0.07 & 0.07 & 0.00 & 0.07 \\
 & Power & Train & 1.00 & 1.00 & 1.00 & 1.00 & 1.00 & 1.00 & 1.00 & 1.00 \\
 &  & Test & 1.00 & 0.93 & 1.00 & 1.00 & 1.00 & 1.00 & 1.00 & 1.00 \\
 & ReverseList & Train & 1.00 & 1.00 & 1.00 & 1.00 & 1.00 & 1.00 & 1.00 & 1.00 \\
 &  & Test & 1.00 & 0.93 & 1.00 & 1.00 & 1.00 & 1.00 & 1.00 & 1.00 \\
 & ReverseNumber & Train & 1.00 & 1.00 & 1.00 & 1.00 & 1.00 & 1.00 & 1.00 & 1.00 \\
 &  & Test & 1.00 & 0.93 & 1.00 & 1.00 & 1.00 & 1.00 & 1.00 & 1.00 \\
 & Sign & Train & 1.00 & 1.00 & 1.00 & 1.00 & 1.00 & 1.00 & 1.00 & 1.00 \\
 &  & Test & 1.00 & 0.93 & 1.00 & 0.87 & 1.00 & 1.00 & 1.00 & 1.00 \\
 & SparseParity & Train & 0.88 & 0.88 & 1.00 & 1.00 & 1.00 & 1.00 & 1.00 & 1.00 \\
 &  & Test & 0.47 & 0.40 & 0.33 & 0.47 & 0.40 & 0.40 & 1.00 & 0.60 \\
 & SumDigits & Train & 1.00 & 1.00 & 1.00 & 1.00 & 1.00 & 1.00 & 1.00 & 1.00 \\
 &  & Test & 1.00 & 0.93 & 1.00 & 0.87 & 1.00 & 1.00 & 1.00 & 1.00 \\
 & TrailingZeros & Train & 1.00 & 1.00 & 1.00 & 1.00 & 1.00 & 1.00 & 1.00 & 1.00 \\
 &  & Test & 0.80 & 0.73 & 0.73 & 0.67 & 0.80 & 0.80 & 0.73 & 0.80 \\
\midrule
Geometry & CheckStraightLine & Train & 1.00 & 1.00 & 1.00 & 1.00 & 1.00 & 1.00 & 1.00 & 1.00 \\
 &  & Test & 0.80 & 0.73 & 0.73 & 1.00 & 0.80 & 0.80 & 0.66 & 0.73 \\
 & MinTimeVisitPoints & Train & 0.50 & 0.50 & 0.50 & 1.00 & 0.50 & 0.50 & 1.00 & 1.00 \\
 &  & Test & 0.07 & 0.00 & 0.13 & 0.00 & 0.20 & 0.20 & 0.06 & 0.13 \\
 & MaxPointsOnLine & Train & 1.00 & 1.00 & 1.00 & 1.00 & 1.00 & 1.00 & 1.00 & 1.00 \\
 &  & Test & 0.80 & 0.73 & 1.00 & 1.00 & 1.00 & 1.00 & 0.93 & 1.00 \\
 & NumberOfBoomerangs & Train & 0.88 & 0.88 & 0.38 & 0.50 & 0.38 & 0.38 & 0.50 & 0.50 \\
 &  & Test & 0.53 & 0.46 & 0.26 & 0.40 & 0.33 & 0.33 & 0.46 & 0.53 \\
 & ValidSquare & Train & 1.00 & 1.00 & 1.00 & 1.00 & 1.00 & 1.00 & 1.00 & 1.00 \\
 &  & Test & 1.00 & 0.93 & 0.93 & 1.00 & 1.00 & 1.00 & 1.00 & 1.00 \\
 & LargestTriangleArea2 & Train & 0.13 & 0.13 & 0.26 & 1.00 & 0.33 & 0.25 & 1.00 & 1.00 \\
 &  & Test & 0.00 & 0.00 & 0.26 & 0.40 & 0.33 & 0.33 & 0.46 & 0.53 \\
 & BoundingBoxArea & Train & 0.00 & 0.00 & 1.00 & 1.00 & 1.00 & 1.00 & 1.00 & 1.00 \\
 &  & Test & 0.00 & 0.00 & 0.93 & 1.00 & 1.00 & 1.00 & 1.00 & 1.00 \\
 & BoundingBoxPerimeter & Train & 0.38 & 0.53 & 1.00 & 1.00 & 1.00 & 1.00 & 1.00 & 1.00 \\
 &  & Test & 0.60 & 0.53 & 0.93 & 1.00 & 1.00 & 1.00 & 1.00 & 1.00 \\
 & ChebyshevDistance & Train & 0.00 & 0.00 & 1.00 & 1.00 & 1.00 & 1.00 & 1.00 & 1.00 \\
 &  & Test & 0.00 & 0.00 & 0.93 & 1.00 & 0.93 & 1.00 & 1.00 & 1.00 \\
 & EuclideanDistanceSquared & Train & 1.00 & 1.00 & 1.00 & 1.00 & 1.00 & 1.00 & 1.00 & 1.00 \\
 &  & Test & 1.00 & 0.93 & 1.00 & 1.00 & 1.00 & 1.00 & 1.00 & 1.00 \\
 & HasDuplicatePoints & Train & 1.00 & 1.00 & 1.00 & 1.00 & 1.00 & 1.00 & 1.00 & 1.00 \\
 &  & Test & 1.00 & 0.93 & 0.86 & 0.87 & 0.93 & 0.93 & 0.93 & 1.00 \\
 & IsAxisAligned & Train & 1.00 & 1.00 & 1.00 & 1.00 & 1.00 & 1.00 & 1.00 & 1.00 \\
 &  & Test & 0.93 & 0.86 & 0.86 & 1.00 & 0.93 & 0.93 & 0.86 & 0.93 \\
 & ManhattanDistance & Train & 1.00 & 1.00 & 1.00 & 1.00 & 1.00 & 1.00 & 1.00 & 1.00 \\
 &  & Test & 1.00 & 0.93 & 1.00 & 1.00 & 1.00 & 1.00 & 1.00 & 1.00 \\
 & MidpointIntegerCheck & Train & 1.00 & 1.00 & 1.00 & 1.00 & 1.00 & 1.00 & 1.00 & 1.00 \\
 &  & Test & 0.80 & 0.73 & 0.73 & 0.67 & 0.80 & 0.80 & 0.73 & 0.80 \\
 & Orientation & Train & 0.75 & 0.75 & 1.00 & 1.00 & 1.00 & 1.00 & 1.00 & 1.00 \\
 &  & Test & 0.40 & 0.33 & 0.20 & 0.40 & 0.27 & 0.27 & 0.46 & 0.53 \\
 & PathReturnsOrigin & Train & 0.88 & 0.88 & 1.00 & 1.00 & 1.00 & 1.00 & 1.00 & 1.00 \\
 &  & Test & 0.93 & 0.86 & 0.93 & 1.00 & 0.93 & 1.00 & 0.93 & 1.00 \\
 & RectangleAreaByDiagonal & Train & 0.00 & 0.00 & 1.00 & 1.00 & 1.00 & 1.00 & 1.00 & 1.00 \\
 &  & Test & 0.00 & 0.00 & 0.93 & 0.87 & 0.93 & 1.00 & 1.00 & 1.00 \\
 & TriangleArea2 & Train & 1.00 & 1.00 & 1.00 & 1.00 & 1.00 & 1.00 & 1.00 & 1.00 \\
 &  & Test & 1.00 & 0.93 & 1.00 & 1.00 & 1.00 & 1.00 & 1.00 & 1.00 \\
\midrule
Multimodal & CellCount & Train & -- & 0.53 & 0.60 & 0.67 & 0.73 & 0.75 & 0.73 & 0.88 \\
 &  & Test & -- & 0.40 & 0.47 & 0.47 & 0.60 & 0.60 & 0.60 & 0.60 \\
 & DiskCount & Train & -- & 0.80 & 0.93 & 0.87 & 1.00 & 1.00 & 1.00 & 1.00 \\
 &  & Test & -- & 0.67 & 0.80 & 0.67 & 1.00 & 1.00 & 1.00 & 1.00 \\
 & MaxRadius & Train & -- & 0.20 & 0.27 & 0.33 & 0.40 & 0.25 & 0.47 & 0.63 \\
 &  & Test & -- & 0.07 & 0.13 & 0.13 & 0.27 & 0.13 & 0.33 & 0.47 \\
 & PeakCount & Train & -- & 0.27 & 0.27 & 0.40 & 0.47 & 0.25 & 0.53 & 0.63 \\
 &  & Test & -- & 0.13 & 0.13 & 0.20 & 0.33 & 0.20 & 0.40 & 0.53 \\
 & TopIndex & Train & -- & 0.20 & 0.20 & 0.27 & 0.27 & 0.25 & 0.27 & 0.50 \\
 &  & Test & -- & 0.07 & 0.07 & 0.07 & 0.13 & 0.00 & 0.13 & 0.20 \\
 & UpperCount & Train & -- & 0.13 & 0.13 & 0.20 & 0.20 & 0.00 & 0.20 & 0.25 \\
 &  & Test & -- & 0.00 & 0.00 & 0.00 & 0.07 & 0.00 & 0.07 & 0.00 \\
 & PolySides & Train & -- & 0.27 & 0.27 & 0.40 & 0.33 & 0.00 & 0.53 & 0.75 \\
 &  & Test & -- & 0.13 & 0.13 & 0.20 & 0.20 & 0.00 & 0.40 & 0.73 \\
 & PieMax & Train & -- & 0.27 & 0.27 & 0.27 & 0.27 & 0.38 & 0.27 & 0.38 \\
 &  & Test & -- & 0.13 & 0.13 & 0.07 & 0.13 & 0.20 & 0.13 & 0.20 \\
 & ArrowDir & Train & -- & 0.20 & 0.20 & 0.27 & 0.20 & 0.38 & 0.20 & 0.63 \\
 &  & Test & -- & 0.07 & 0.07 & 0.07 & 0.07 & 0.00 & 0.07 & 0.07 \\
 & RadarMax & Train & -- & 0.20 & 0.27 & 0.27 & 0.20 & 0.38 & 0.27 & 0.50 \\
 &  & Test & -- & 0.07 & 0.13 & 0.07 & 0.07 & 0.07 & 0.13 & 0.27 \\
\end{longtable}
\endgroup

\section{Hyperparameter Impact} \label{app:sensitivity}

We further investigate the sensitivity of DIO-Agent to two key hyperparameters: the number of independent evolutionary islands and the total number of search iterations. These parameters represent the balance between exploration breadth and search depth.
\begin{figure*}[h]
    \centering
    \includegraphics[width=\textwidth]{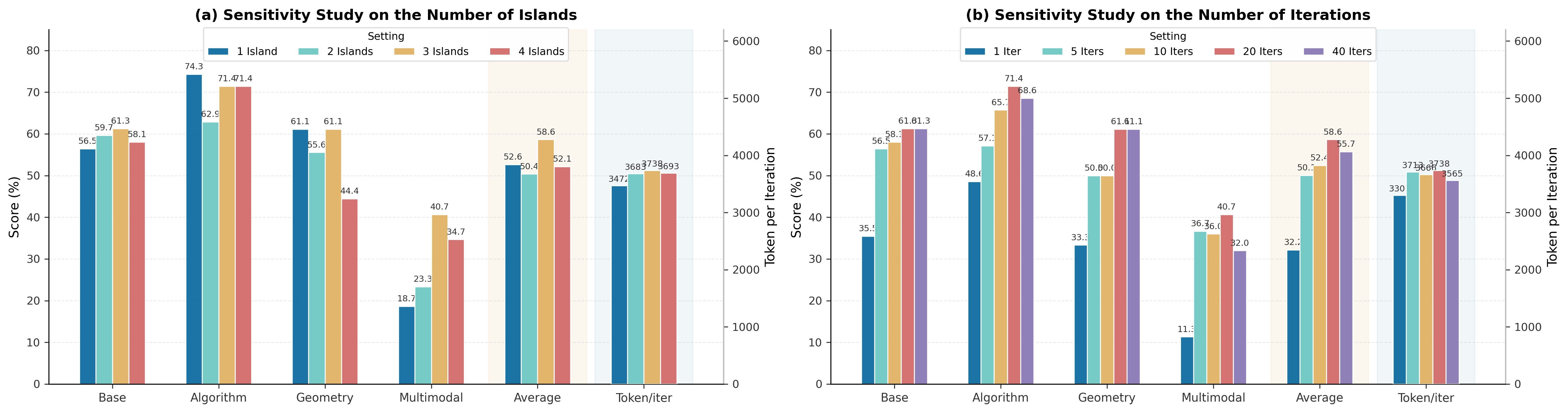}
    \caption{
    Sensitivity studies on different hyperparameters.
    (a) Sensitivity study on the number of islands.
    (b) Sensitivity study on the number of iterations.
    }
    \label{fig:sensitivity_studies}
\end{figure*}

\noindent \textbf{Impact of Island Count.} As shown in Figure~\ref{fig:sensitivity_studies}(a), the average performance peaks when using 3 islands, achieving a pass rate of 64.61\%. While a single island provides surprisingly strong results in the Algorithm level (74.29\%), it struggles with the Base level, likely due to a lack of diverse candidate rules during the early stages of discovery. Increasing the count to 4 islands leads to a performance regression, particularly in the Geometry level (dropping to 44.44\%). This suggests that while parallel islands encourage exploration and prevent premature convergence, an excessive number of islands may dilute the quality of the population or lead to resource fragmentation within the fixed iteration budget, thereby hindering the fine-tuning of complex geometric logic.

\noindent \textbf{Impact of Search Iterations. }Figure~\ref{fig:sensitivity_studies}(b) illustrates the relationship between the number of iterations and the discovery success rate. Performance scales positively with search depth up to 20 iterations, where the average pass rate reaches its maximum of 64.61\%. The improvement is most significant between 5 and 20 iterations, particularly in the Algorithm and Geometry levels, which require multiple refinement steps to identify non-obvious algorithmic patterns. Interestingly, increasing the depth further to 40 iterations does not yield additional gains; instead, we observe a slight dip in the Algorithm level performance (68.57\%). This plateau indicates that 20 iterations provide a sufficient search horizon for the current benchmark, and further expansion may lead to diminishing returns or marginal overfitting on visible examples that does not translate to held-out test cases. Notably, the token cost remains relatively stable across these variations, suggesting that the complexity of each iteration is well-contained by our framework's structured design. 

\section{Detailed Ablation Results}
\label{Detailed Ablation Results}
To understand the individual contribution of each component, we conduct an ablation study across all difficulty levels. The results are shown in Table \ref{tab:ablation_detail}.
The most critical component is Curriculum-wise Evolution (CE). Removing it results in the largest performance drop, with the average pass rate falling from 58.63\% to 51.33\%. This decline is particularly pronounced in the Algorithm level (from 71.43\% to 62.86\%). Furthermore, its absence increases the token cost per iteration to 4244.31, suggesting that the curriculum not only improves accuracy but also makes the search more efficient by pruning unproductive paths early.
TPP also plays a vital role, particularly in specialized domains. Its removal leads to a substantial decrease in Geometry performance (from 61.11\% to 50.00\%) and a noticeable drop in Algorithm (from 71.43\% to 68.57\%), indicating that structured complexity ordering is crucial for tasks involving non-trivial algorithmic structures.
Error-Grounded Feedback (EF) serves as a critical refinement tool. While its impact on the Algorithm level is negligible, its absence causes a sharp decline in Geometry (from 61.11\% to 50.00\%) and Multimodal (from 40.67\% to 37.33\%), suggesting that error traces are instrumental in correcting precise logical errors common in these domains.

\begin{table}[h]
\centering
\caption{Ablation Study, where each module is removed separately.}
\small{
\begin{tabular}{lcccccc}
\toprule
\multirow{2}{*}{Method} & \multicolumn{6}{c}{IO2CodeBench}\\
\cmidrule(l){2-7} & \multicolumn{1}{l}{Base} & \multicolumn{1}{l}{Algorithm} & \multicolumn{1}{l}{Geometry}& \multicolumn{1}{l}{Multimodal} & \multicolumn{1}{l}{Average} & \multicolumn{1}{l}{Token/iter} \\
\midrule
DIO-Agent          & \textbf{61.29}                    & \textbf{71.43}                     & 61.11       & 40.67                 & 58.63                      & 3738.31                      \\
\midrule
w/o EF  & 59.68                    & 71.43                     & 50.00             &    37.33       &     54.61                   & 3837.61                      \\
w/o TPP & 61.29                    & 68.57                     & 50.00             &    35.33       &        53.80                & 3369.48                      \\
w/o CE           & 56.45                    & 62.86                     & 50.00                        &    36.00           &  51.33      & 4244.31                      \\
\bottomrule
\end{tabular}}
\label{tab:ablation_detail}
\end{table}

\section{Detailed Behavioral Analysis}
\label{app:behavioral_analysis}
This appendix provides additional behavioral analyses of DIO-Agent on non-multimodal tasks. We exclude the Multimodal tier from these analyses because multimodal tasks involve an auxiliary LLM for image-to-value extraction, making program complexity, mutation attribution, and failure trajectories less directly comparable to purely code-based tasks.

Table~\ref{tab:dio_complexity_trajectory_success} analyzes how the complexity of the best program evolves across curriculum stages. Successful runs are often associated with either stable complexity or monotonic simplification. In particular, the monotone-down pattern achieves the highest held-out success rate, suggesting that DIO-Agent frequently discovers a more general rule by first proposing a workable structure and then compressing unnecessary logic. In contrast, monotone-up trajectories have a much lower success rate, which is consistent with the failure mode of accumulating patches to satisfy visible examples rather than identifying the underlying transformation.

\begin{table}[h]
\centering
\caption{Complexity trajectories and held-out success.}
\label{tab:dio_complexity_trajectory_success}
\resizebox{\textwidth}{!}{
\begin{tabular}{lrrrr}
\toprule
Trajectory type & Runs & Held-out successes & Success rate & Interpretation \\
\midrule
Stable & 37 & 31 & 83.8\% & Rule found early; little structural change \\
Monotone down & 18 & 17 & 94.4\% & Continuous simplification \\
Hump: up then down & 18 & 12 & 66.7\% & Explore, then compress \\
Monotone up & 34 & 9 & 26.5\% & Patch accumulation / overfitting \\
Valley: down then up & 4 & 3 & 75.0\% & Rare mixed simplification pattern \\
Mixed & 4 & 1 & 25.0\% & No consistent direction \\
\midrule
Total & 115 & 73 & 63.5\% & \\
\bottomrule
\end{tabular}}
\end{table}

Table~\ref{tab:dio_difficulty_adjusted_trajectory} further controls for task difficulty. Task difficulty is estimated from non-hybrid baselines and is not computed from the focal DIO-Agent run, so the residual reflects whether a trajectory type outperforms tasks of comparable estimated difficulty. The positive residual for monotone-down trajectories shows that their high success rate is not merely a consequence of easier tasks. Conversely, monotone-up trajectories remain strongly negative after difficulty adjustment, reinforcing the interpretation that increasing complexity across stages is a warning sign for overfitting or patch accumulation.

\begin{table}[h]
\centering
\small
\caption{Difficulty-adjusted trajectory analysis.}
\label{tab:dio_difficulty_adjusted_trajectory}
\begin{tabular}{lrrrr}
\toprule
Trajectory type & Hybrid runs & Success rate & Avg. task difficulty & Success residual \\
\midrule
Monotone down & 173 & 90.8\% & 0.750 & +0.212 \\
Stable & 203 & 69.0\% & 0.793 & -0.029 \\
Hump: up then down & 148 & 61.5\% & 0.541 & +0.047 \\
Monotone up & 207 & 24.2\% & 0.357 & -0.228 \\
Valley: down then up & 44 & 65.9\% & 0.389 & +0.216 \\
Mixed & 29 & 51.7\% & 0.363 & -0.002 \\
\bottomrule
\end{tabular}
\end{table}

Table~\ref{tab:dio_visible_overfit} distinguishes visible-set fitting from true held-out generalization. Among the held-out failures, many programs achieve perfect accuracy on the final visible curriculum stage but still fail on hidden test cases. These curriculum-perfect failures are substantially longer than held-out-successful programs, indicating that perfect visible accuracy alone is not a reliable indicator of rule discovery. Instead, excessive program length often reflects memorization or case-specific branching that does not transfer to unseen inputs.

\begin{table}[h!]
\centering
\caption{Visible-stage success versus held-out generalization.}
\label{tab:dio_visible_overfit}
\resizebox{\textwidth}{!}{
\begin{tabular}{lrrrr}
\toprule
Group & Runs & Visible stage-4 accuracy & Held-out outcome & Avg. code length \\
\midrule
Held-out-successful runs & 73 & -- & Success & 249.8 chars \\
Held-out failures & 42 & -- & Failure & -- \\
Stage-4 curriculum-perfect but held-out-failed & 36 & 1.0 & Failure & 833.9 chars \\
\bottomrule
\end{tabular}}
\end{table}

\begin{table}[h!]
\centering
\small
\caption{Overfitting failures by task family.}
\label{tab:dio_overfit_by_family}
\begin{tabular}{lrrrr}
\toprule
Task family & Tasks & Overfit failures & Overfit rate & Avg. failed length \\
\midrule
Arithmetic & 15 & 9 & 60.0\% & 1092.2 chars \\
Core & 12 & 6 & 50.0\% & 806.5 chars \\
LeetCode\_Geo & 18 & 7 & 38.9\% & 246.7 chars \\
Extra & 35 & 11 & 31.4\% & 758.5 chars \\
BitParity & 13 & 2 & 15.4\% & 2268.0 chars \\
Sequence & 18 & 1 & 5.6\% & 743.0 chars \\
Newton & 4 & 0 & 0.0\% & -- \\
\bottomrule
\end{tabular}
\end{table}

Table~\ref{tab:dio_overfit_by_family} breaks down the curriculum-perfect but held-out-failed cases by task family. Arithmetic and Core tasks show the highest overfitting rates, suggesting that these families are especially prone to visible-set shortcuts. This is plausible because arithmetic tasks often admit many partial rules that match a small number of examples but fail under broader coverage. In contrast, Sequence and Newton tasks show much lower overfitting rates, indicating that visible examples in these families may provide stronger constraints on the underlying rule.

\begin{figure}[h!]
    \centering
    \includegraphics[width=0.73\linewidth]{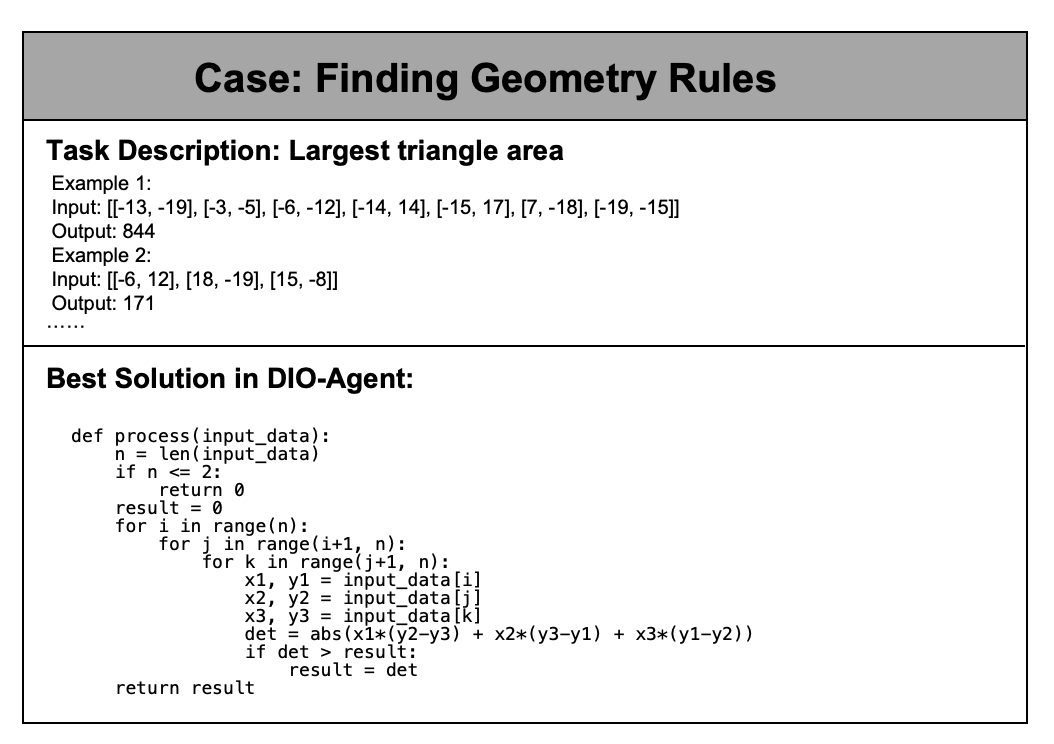}
    \caption{LLM manages to discover the geometry rule behind the I/O examples}
    \label{fig:case1}
\end{figure}
\begin{figure}[h!]
    \centering
    \includegraphics[width=0.73\linewidth]{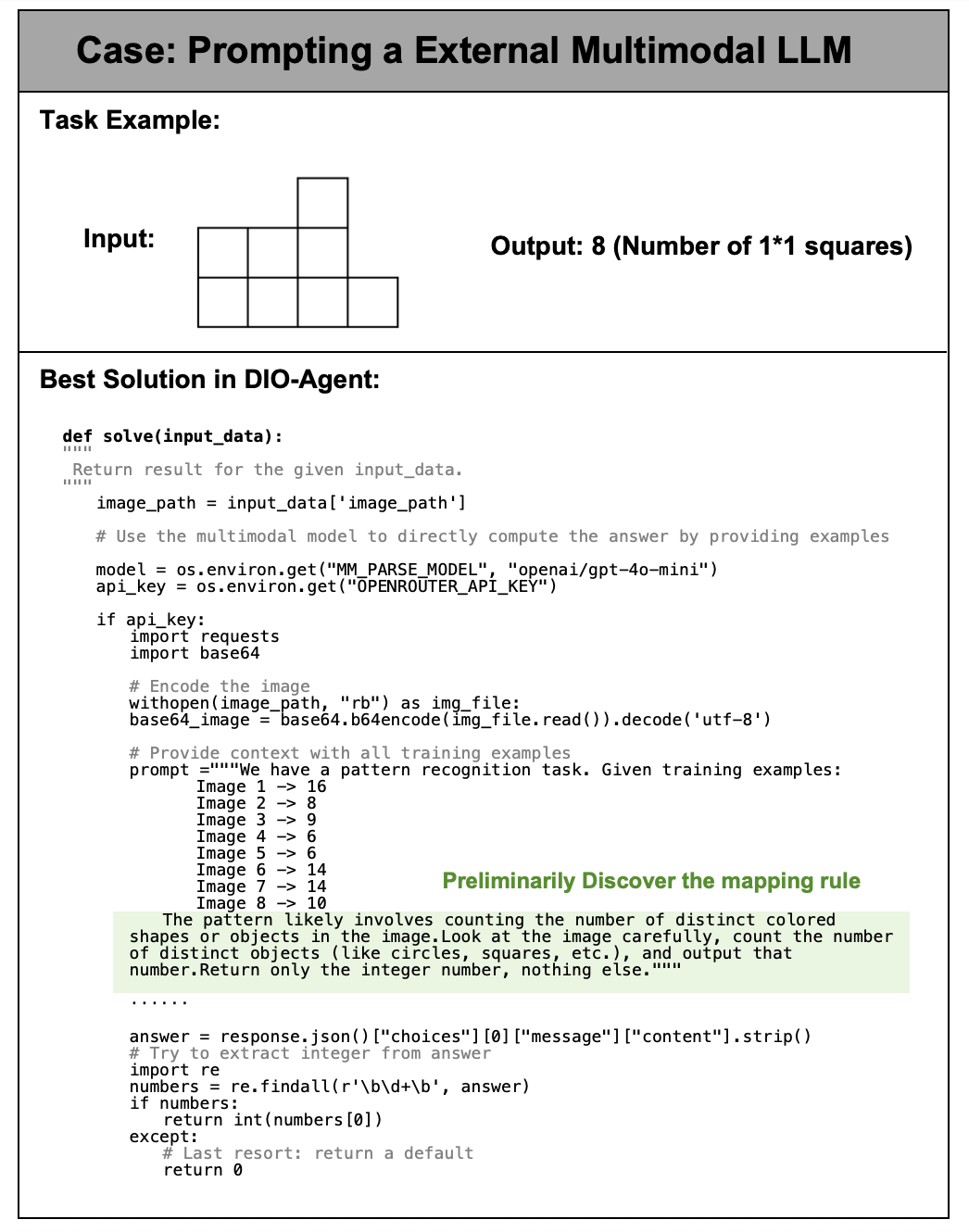}
    \caption{LLM utilizes an external multimodal LLM as a tool to solve Multimodal problems}
    \label{fig:case2}
\end{figure}
\begin{figure}[h]
    \centering
    \includegraphics[width=0.73\linewidth]{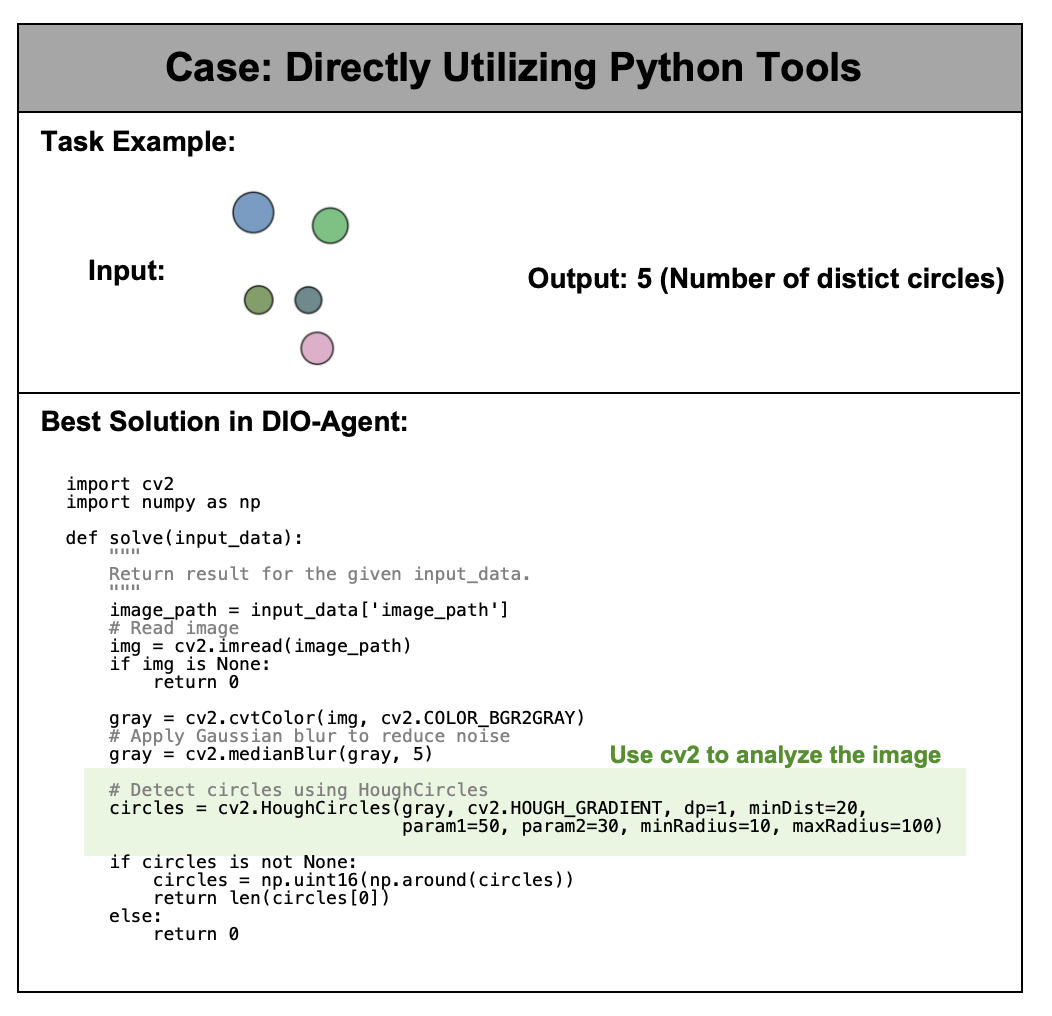}
    \caption{LLM directly utilizes python packages as tools to solve Multimodal problems}
    \label{fig:case3}
\end{figure}

\section{Case Study}
\label{sec:case study}
In this section, we study the performance of DIO-Agent across three specific problems. As shown in Figure \ref{fig:case1}, the first case involves a geometry task. The input consists of an indefinite number of point coordinates (greater than 3), and the required output is the area of the largest triangle formed by these points. Under our Curriculum-wise Evolution approach, the model successfully discovered the underlying relationship between the inputs and outputs. The second case explores a multimodal scenario, illustrated in Figure \ref{fig:case2}. The input is a grid image, and the output corresponds to the total number of $1 \times 1$ squares within the image. During the solution process, the model opted to call our provided multimodal model as a tool to extract specific visual information, leading to a successful answer. The third case is also a multimodal scenario, shown in Figure \ref{fig:case3}. The input image contains circles of various colors (where the color variations serve as distractors), and the output is the total count of circles. In this instance, the LLM chose to directly utilize the cv2 package, calling specific functions for information extraction to achieve the final solution. These experiments demonstrate that DIO-Agent can learn to diversify its problem-solving strategies based on the specific requirements of different task types.

\begin{figure} [h!]
    \centering
    \includegraphics[width=0.9\linewidth]{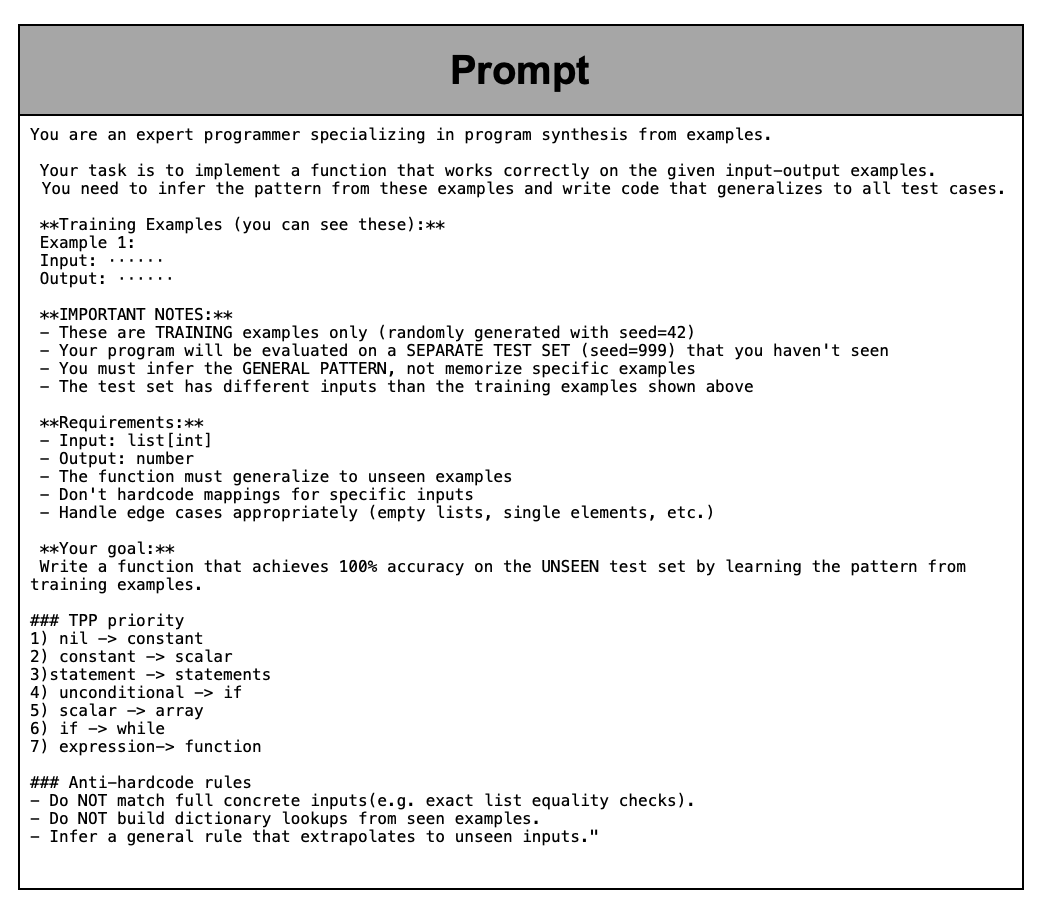}
    \caption{Prompts in DIO-Agent.}
    \label{fig: prompt}
\end{figure}

\section{Limitation} \label{appendix:limitations}

Our work has two main limitations. First, although the proposed IO2CodeBench covers multiple difficulty levels ranging from basic data operations to complex algorithmic reasoning (e.g., geometric computation and multimodal tasks), the landscape of code generation includes numerous specialized fields yet to be fully explored. Expanding the domain coverage would enable a more comprehensive evaluation of different methods, and this constitutes an important direction for our future work.
Second, the current experimental setup uniformly provides fixed visible I/O examples per task for search. However, in real-world applications, the number of available I/O examples may vary considerably, and different examples can differ significantly in their informativeness for revealing the underlying rule. Certain well-chosen boundary cases may be far more discriminative than a large number of routine examples. The impact of the quantity, diversity, and informational coverage of I/O examples on search effectiveness warrants further systematic investigation.

\section{IO2CodeBench} \label{appdendix:details}
This appendix describes the construction of the IO2CodeBench used in our experiments. The benchmark is programmatically generated from deterministic oracle functions, therefore every example pair is produced automatically rather than annotated by humans.

\subsection{Benchmark Composition}

We establish our IO2CodeBench for experiments, informed by \citep{OAIBB}. The IO2CodeBench includes Base, Algorithm, Geometry, and Multi-modal levels. Each task corresponds to a deterministic target program defined in a task registry and spans a different transformation pattern over structured inputs. At a high level, the benchmark covers cumulative arithmetic operations, memory-style sequence transductions, parity and bitwise reasoning, multi-stream operations such as base-$k$ addition and binary dot products, and several small symbolic reasoning tasks.

Each task is defined by a ground-truth function $f^*$ together with a task-specific input domain. For the core benchmark, the domain type is one of five categories: integer sequences, bit sequences, paired integer sequences, paired bit sequences, or floating-point sequences. The labels are always generated by executing the oracle,
\begin{equation}
y = f^*(x),
\end{equation}
so each task is fully specified by executable semantics rather than by natural-language annotation.

\subsection{Example Generation}

For the core sequence tasks, inputs are sampled automatically according to the task type and range constraints stored in the registry. Sequence lengths are drawn uniformly between 3 and 10. Integer-valued tasks sample each element from a task-specific bounded interval, bit-valued tasks sample each element from $\{0,1\}$, floating-point tasks sample from a bounded real interval, and two-stream tasks generate two aligned sequences of the same length. The target output is then obtained by directly applying the hidden oracle function to the sampled input.

For the extended tasks, we use task-specific generators. These include bounded integer problems, strings, integer lists, reverse-Polish-expression sequences, bit-vector problems, and simple geometry tasks over points and paths. To improve coverage, the generator first inserts a small set of canonical edge cases and then fills the remaining examples with random samples from the corresponding domain. This produces a benchmark that is still automatically generated, but less homogeneous than simple iid random sampling.

\subsection{Train--Test Split Construction}

By default, we construct 8 visible training examples and 15 hidden test examples per task, where each set ensures the 100\% test coverage of the task. The training set is generated with random seed 42 and the test set with random seed 999. This separation is hard-coded in the generation pipeline and is also stated explicitly in the prompt shown to the model.

The split procedure is designed to avoid leakage at the input level. For each task, we first sample an oversized candidate pool for training and testing, deduplicate examples by input, and then remove any test input that also appears in the training pool. If the unique pool is still too small, the generator performs an additional top-up with backup seeds.
Thus, the benchmark enforces train--test input disjointness whenever the task domain permits it.

Formally, for task $\tau$ we construct a visible set $\mathcal{D}^{(\tau)}_{\mathrm{train}}$ and a hidden set $\mathcal{D}^{(\tau)}_{\mathrm{test}}$ such that
\begin{equation}
\mathcal{D}^{(\tau)}_{\mathrm{train}}=\{(x_i,f^*_\tau(x_i))\}_{i=1}^{n_{train}}, \qquad
\mathcal{D}^{(\tau)}_{\mathrm{test}}=\{(x_j,f^*_\tau(x_j))\}_{j=1}^{n_{test}},
\end{equation}
with disjoint inputs whenever possible. During search, only $\mathcal{D}^{(\tau)}_{\mathrm{train}}$ is visible to the synthesizer; $\mathcal{D}^{(\tau)}_{\mathrm{test}}$ is never revealed to the model.

\subsection{Evaluation Protocol}

Evaluation is execution-based. Given a synthesized program $\hat{p}$, the evaluator imports the generated Python file, locates the task-specific function, and executes it on every hidden test example. A prediction is marked correct if it exactly matches the expected output. For structured outputs such as lists, comparison is performed recursively element by element; for floating-point values, we allow a small numerical tolerance of $10^{-6}$.

For a task with test set $\mathcal{D}_{\mathrm{test}}$, the final task-level accuracy is
\begin{equation}
\mathrm{Acc}_{\mathrm{test}}(\hat{p})=
\frac{1}{|\mathcal{D}_{\mathrm{test}}|}
\sum_{(x,y)\in\mathcal{D}_{\mathrm{test}}}\mathbf{1}[\hat{p}(x)=y].
\end{equation}
The evaluator also records the number of correct cases, total cases, runtime, and up to three concrete failure examples for diagnosis. In the final hidden evaluation used for benchmark reporting, the selection score equals this held-out accuracy.

At the benchmark level, we report both average hidden-set accuracy and \emph{task success rate}, where a task is counted as solved only if the synthesized program achieves $\mathrm{Acc}_{\mathrm{test}}=1.0$. This exact-match criterion is important for PBE, since partially correct programs may fit many visible examples while still failing to recover the underlying rule.

\subsection{Search-Time vs. Final Evaluation}

One final distinction is important. During search, the synthesizer is never optimized directly on the hidden test set. In the baseline setting, search uses a training-only evaluator built from the 8 visible examples. In DIO-Agent, search further decomposes these visible examples into curriculum stages and uses replay examples from earlier stages for regression checking. However, in all cases, the numbers reported in the main tables are obtained from the final hidden evaluator on the 15-example test split. This keeps optimization and reporting cleanly separated and makes the benchmark a true generalization test rather than a memorization exercise.

\section{Experimental Setup}
\label{sec: setup}

\noindent \textbf{Benchmark.}
Our proposed evaluation benchmark comprises 125 IO2Code tasks organized into four difficulty levels designed to test progressively more complex pattern discovery capabilities. The \textbf{Base} level contains 62 tasks involving scalar arithmetic, list operations, and standard map/filter/reduce patterns. The underlying rule here can typically be expressed in one to three lines of Python. Examples include reversing a list, computing a running sum, or applying element-wise arithmetic. The \textbf{Algorithm} level contains 35 tasks requiring algorithmic reasoning, dynamic programming, or multi-step transformations. The mapping from input to output here involves non-obvious algorithmic structure such as sorting by a derived key, computing edit distances, or implementing combinatorial logic. The \textbf{Geometry} level contains 18 tasks drawn from geometric computation problems, including area calculations, intersection detection, and coordinate transformations. In addition, the \textbf{Multi-modal} level contains 10 tasks where the inputs are images and the outputs correspond to specific numerical properties within the images (such as the number of shapes or spatial relationships). The model is provided with an external multi-modal LLM tool, which it can freely call to extract information from the images for further computational processing~\citep{qwen2026qwen35}.
Each task provides 8 visible I/O examples used during search and 15 held-out test cases used exclusively for evaluation. The level structure is designed to separate tasks susceptible to ceiling effects (Base) from those where structured search provides meaningful differentiation (Algorithm, Geometry, Multi-modal). The prompts are shown in Figure \ref{fig: prompt}. In the Autonomous Discovery setting, the model initially constructs 2 I/O examples. Once the evolved code successfully passes all existing examples, the model is permitted to generate additional I/O pairs, increasing the sample count by 2 in each subsequent step. We set the maximum number of iterations to 50 and implement an early stopping strategy if the model consistently satisfies all current examples for five consecutive rounds.

\noindent \textbf{Evaluation Metric.} 
To reflect the differing nature of the tasks, we employ a tiered scoring system. For the Base, Algorithm, and Geometry levels, we report the level Pass Rate~\citep{chen2021evaluatinglargelanguagemodels,austin2021programsynthesislargelanguage}. A task is considered as solved only if the generated program $\hat{p}$ correctly satisfies $\hat{p}(x_i) = y_i$ for all 15 held-out test cases. The metric for these levels is the percentage of tasks solved within each category. Conversely, for the Multimodal level, we utilize a more granular Mean Sample Pass Ratio. For each of the 10 multimodal tasks, we calculate the ratio of held-out test cases passed by the model. The final metric for this level is the average of these ratios across all 10 tasks. This approach accounts for the inherent complexity of image-to-value extraction, where partial success across test cases provides a more nuanced measure of model capability. As a secondary metric, we report the Computational Efficiency via the token cost per iteration. This is decomposed into the mean prompt tokens and mean completion tokens across all LLM calls. 

\noindent \textbf{Baselines.} We compare DIO-Agent against seven baselines spanning the spectrum from classical PBE methods to modern LLM-based evolution-agent baselines.
Among these, Direct represents a standard approach where the LLM is prompted to generate code directly based on visible examples~\citep{brown2020languagemodelsfewshotlearners,chen2021evaluatinglargelanguagemodels}. Meanwhile, PBE and E-PBE are methods rooted in the Programming-by-Example paradigm within the software engineering domain~\citep{gulwani2016programming,verbruggen2025execution}. TDD means TDD-guided Generation iteratively. Moreover, FunSearch, AlphaEvolve, and CodeEvolve represent outstanding evolutionary algorithms tailored for the field of code generation~\citep{romera2024mathematical,novikov2025alphaevolve,assumpccao2025codeevolve}. We conduct all experiments here with DeepSeek-v3.2 as the base LLM to ensure fair comparison across methods~\citep{deepseekai2025deepseekv32pushingfrontieropen}.

\end{document}